%% file: thesis_main.tex
\documentclass[11pt,
  a4paper,
  parskip=half, 
  BCOR=10mm, 
  english,
  ]{scrbook}
\usepackage[english]{babel} 
\usepackage{amsmath, amsfonts,amssymb}
\usepackage[algosection ,vlined, ruled, algo2e, linesnumbered]{algorithm2e} 
\usepackage[normalem]{ulem}

\usepackage{caption}
\usepackage{mathtools}
\newcommand{\expnumber}[2]{{#1}\mathrm{e}{#2}}

\usepackage[square,numbers]{natbib}
\usepackage{titling}
\title{Dynamical Audio-Visual Navigation: Catching Unheard Moving Sound Sources in Unmapped 3D Environments}
\author{Abdelrahman Younes}

\newcommand{\firstexaminer}{Prof.~Dr.~Abhinav Valada}
\newcommand{\secondexaminer}{Prof.~Dr.~Joschka Boedcker}
\newcommand{\advisers}{ 
Dr. Tim Welschehold, Daniel Honerkamp}

\input{setup}

\begin{document}
\newcolumntype{P}[1]{>{\centering\arraybackslash}p{#1}}
    \pagestyle{empty} 
    \hypersetup{pageanchor=false}

    \input{chapters/0_0-titlepage_en}

    \pagestyle{plain} 
    \frontmatter  
    \input{chapters/0_1-declaration}
    \input{chapters/dedication}

    \input{chapters/0_2-abstract}
    \input{chapters/0_3-acknowledgments}
    \tableofcontents
    \listoffigures
    \listoftables
    \listofalgorithms
    \hypersetup{pageanchor=true}  

    \mainmatter  
    \input{chapters/1-introduction}
    \input{chapters/2-background}
    \input{chapters/3-relatedwork}
    \input{chapters/4-datasets}
    \input{chapters/5-approach}
    \input{chapters/6-experiments}
    \input{chapters/8-conclusions}

    \cleardoublepage
    \phantomsection
    \addcontentsline{toc}{chapter}{Bibliography}
    \bibliography{bib/topic1}
    \newpage
    \thispagestyle{empty}
    \mbox{}

\end{document}

%% file: setup.tex
    \usepackage[T1]{fontenc}  
    \usepackage[utf8x]{inputenc}
    \usepackage{scrhack}

    \usepackage[labelfont=bf, labelsep=colon, format=hang, textfont=singlespacing]{caption}
    \usepackage{chngcntr}  
    \counterwithout{equation}{chapter}
    \counterwithout{figure}{chapter}
    \counterwithout{table}{chapter}

    \setkomafont{chapter}{\normalfont\bfseries\huge}

    \usepackage{setspace}  
    \usepackage[dvipsnames]{xcolor}  
    \usepackage{array}  
    \usepackage{graphicx}  
    \usepackage{subfig}  
    \usepackage{amsmath}  
    \usepackage{amsthm}   
    \usepackage{amsfonts} 
    \usepackage{calc}  
    \usepackage[unicode=true,bookmarks=true,bookmarksnumbered=true,
                bookmarksopen=true,bookmarksopenlevel=1,breaklinks=false,
                pdfborder={0 0 0},backref=false,colorlinks=false]{hyperref}
    \usepackage{etoolbox} 

    \usepackage{rotating}
    \usepackage{tikz}
   \usepackage{algorithm,algpseudocode} 
    \usepackage{bm}  
    \usepackage{tikz}
    \usetikzlibrary{positioning}  
    
    \usepackage{ifthen}

    \usepackage{booktabs}

    \usepackage{lipsum}

    \renewcommand{\eqref}[1]{Equation~(\ref{#1})}

    \definecolor{darkgreen}{rgb}{0.0, 0.5, 0.0}
    \definecolor{UniBlue}{RGB}{0, 74, 153}
    \definecolor{UniRed}{RGB}{193, 0, 42}
    \definecolor{UniGrey}{RGB}{154, 155, 156}

    \let\mySection\section\renewcommand{\section}{\suppressfloats[t]\mySection}
    \let\mySubSection\subsection\renewcommand{\subsection}{\suppressfloats[t]\mySubSection}

    \hypersetup{pdftitle={\thetitle},
                pdfauthor={\theauthor},
                pdfsubject={Undergraduate thesis at the Albert Ludwig University of Freiburg},
                pdfkeywords={deep learning, awesome algorithm,  undergraduate thesis},
                pdfpagelayout=OneColumn, pdfnewwindow=true, pdfstartview=XYZ, plainpages=false}

    \tikzset{>=latex}  

    \algnewcommand\algorithmicforeach{\textbf{foreach}}
    \algdef{S}[FOR]{ForEach}[1]{\algorithmicforeach\ #1\ \algorithmicdo}

	\usepackage{ifthen} 
	\newcounter{todos}
	\newcounter{extends}
	\newcounter{drafts}
    \usepackage[binary-units=true]{siunitx}
    \usepackage{romannum}
	\DeclareSIUnit{\million}{\text{M}}
    \DeclareSIUnit{\billion}{\text{B}}

%% file: chapters/0_0-titlepage_en.tex
\begin{titlepage}
\begin{center}

\newcommand{\HorizontalLine}{\rule{\linewidth}{0.3mm}}

{\Large Master's Thesis}\\[1.3cm]

\HorizontalLine \\[0.4cm]
{ \huge \bfseries \thetitle }
\HorizontalLine \\[1.5cm]

{\Huge \theauthor} \\[2cm]

\begin{tabular}[hc]{>{\huge}l >{\huge}l}
  Examiner: & \firstexaminer \\[0.3cm]
  Advisers: & Dr. Tim Welschehold \\[0.3cm]
            & Daniel Honerkamp \\[1.2cm]
\end{tabular}
\vfill  

\Large {
    Albert-Ludwigs-University Freiburg\\
    Faculty of Engineering\\
    Department of Computer Science\\
    Robot Learning Lab\\[1cm]

    December 21\textsuperscript{st}, 2021\\
}
\end{center}
\end{titlepage}

\thispagestyle{empty}
\ \vfill \ \\  
\
\textbf{Writing Period}            \smallskip{} \\
25.\,06.\,2021 -- 21.\,12.\,2021   \bigskip{} \\
\
\textbf{Examiner}                  \smallskip{} \\
\firstexaminer                     \bigskip{} \\
\
\ifdef{\secondexaminer}
	{
	\textbf{Second Examiner}       \smallskip{} \\
	\secondexaminer                \bigskip{} \\
	\
	}
	{
	}
\textbf{Advisers}                  \smallskip{} \\
\advisers 

%% file: chapters/0_1-declaration.tex

\chapter*{Declaration}

I hereby declare, that I am the sole author and composer of my thesis and that no other sources or learning aids, other than those listed, have been used. Furthermore, I declare that I have acknowledged the work of others by providing detailed references of said work.  \newline
I hereby also declare, that my Thesis has not been prepared for another examination
or assignment, either wholly or excerpts thereof.
\\[3\normalbaselineskip]
\begin{tabular}{p{\textwidth/2} l}
  \rule{\textwidth/3}{0.4pt}   &   \rule{\textwidth/3}{0.4pt} \\
  Place, Date                  &   Signature
\end{tabular}

%% file: chapters/dedication.tex
\clearpage
\begin{center}
    \thispagestyle{empty}
    \vspace*{\fill}
    \textit{To the memory of my father,}
    \vspace*{\fill}
\end{center}
\clearpage

%% file: chapters/0_2-abstract.tex
\chapter*{Abstract}
Recent work on audio-visual navigation targets a single static sound in noise-free audio environments and struggles to generalize to unheard sounds. We introduce the novel dynamic audio-visual navigation benchmark in which an embodied AI agent must catch a  moving sound source in an unmapped environment in the presence of distractors and noisy sounds. We propose an end-to-end reinforcement learning approach that relies on a multi-modal architecture that fuses the spatial audio-visual information from a binaural audio signal and spatial occupancy maps to encode the features needed to learn a robust navigation policy for our new complex task settings. We demonstrate that our approach outperforms the current state-of-the-art with better generalization to unheard sounds and better robustness to noisy scenarios on the two challenging 3D scanned real-world datasets Replica and Matterport3D, for the static and dynamic audio-visual navigation benchmarks. Our novel benchmark will be made available at \url{http://dav-nav.cs.uni-freiburg.de}.

%% file: chapters/0_3-acknowledgments.tex
\chapter*{Acknowledgments}
First and foremost, I would like to thank Prof. Dr. Abhinav Valada for giving me this opportunity to work on this exciting topic at the Robot Learning lab. Prof. Valada is one of the game-changers in the Robotics and Machine Learning fields. I always look at his work and career as a source of inspiration. Second, I would like to express my gratitude and appreciation to my advisers Dr. Tim Welschehold and Daniel Honerkamp. I am thankful for the time they spent and the effort they made to supervise me in this project. They provided me with the freedom to pursue my goals in this thesis and helped me to reach them. Our meetings and discussions helped me learn more and widened my thinking horizon. This work would not be possible without their guidance and constant support.  I am also thankful for their eagerness to help me even beyond the thesis scope. Third, I would like to thank my best friend Ola for always being there for me and for her unconditional support. I also want to thank my good friend Salem for encouraging and supporting me and being there for me in the hard times. Last, I sincerely would like to express my appreciation, gratitude, and gratefulness for my mother, aunt, and sisters for being my backbone in this life. Their absolute love and endless support are what keep me going in my life.

%% file: chapters/1-introduction.tex
\chapter{Introduction}
\label{chap:introduction}
Navigation towards acoustic events is part of our daily life. We navigate towards a faucet in the kitchen or bathroom if we hear a water drop, and we know that someone needs help in another room or elevator if we hear screaming or crying sounds. Sound is notably the first signal that most of the devices around us use to inform us that it needs our attention, such as a fire alarm, a dishwasher, a microwave, a ringing telephone, a doorbell, a kettle, and other appliances.  Despite that, most current work focuses on vision only for navigation \cite{gupta2017cognitive,chen2019behavioral}, leaving behind a vital sensory signal like sound, which can reveal information exceeding the capabilities of visual sensors like the presence of a ringing phone in another room as the sound signal can transfer through walls \cite{valada2017deep}. This pivotal sensory signal provides blind people spatial navigation capability comparable to sighted people \cite{fortin2008wayfinding}.

Recently, researchers have tried to use this vital signal in various tasks such as audio-visual navigation, in which the agent has to navigate towards the location of a sound-emitting source relying on audio and visual signals only \cite{chen2020soundspaces,gan2020look}, semantic audio-visual navigation \cite{chen2021semantic}, in which the agent is required to understand the semantics of the sound emitting-source to be able to navigate to the sound source even after the sound stops. The researchers have further introduced non AudioGoal tasks like active audio-visual source separation \cite{majumder2021move2hear}, where the agent is asked to navigate in environments to separate the input sound from others sounds in the environment, audio-visual dereverberation\cite{chen2021learning}, curiosity-based exploration via audio-visual association \cite{valverde2021there} and audio-visual floor plan reconstruction~\cite{purushwalkam2020audio}.

In this work, we tackle the audio-visual navigation task, in which prior research  \cite{chen2020soundspaces, chen2020learning} has mainly focused on tackling the task of navigating towards a sound-emitting source in noise-free audio environments with a single static sound source. The previously mentioned work \cite{chen2020soundspaces, chen2020learning} suffers to generalize to unheard sounds, performs poorly in the presence of distractors and noisy audio observations, and does not cover the scenarios with moving sound sources. To overcome these drawbacks, we first introduce the novel dynamic audio-visual navigation benchmark with a moving sound source. This benchmark extends the range of scenarios to include auditory events with a dynamic sound source, such as a robot following a person issuing commands or babysitting a toddler, following pets or people in the house. This novel benchmark strongly increases the complexity of the task as the agent needs to learn to update its memory more frequently since previous observations no longer represent the current state of the environment. In addition to that, the agent has to proactively reason about the trajectory of the moving sound source to catch it efficiently. 
\\
Our second contribution to address the previously mentioned drawbacks is increasing the complexity of both static and dynamic audio-visual navigation benchmarks by carefully integrating designed complex audio scenarios with distractors as well as introducing noisy and augmented sounds in training to increase the robustness of our agent against this category of scenarios. We demonstrate that these complex audio scenarios benefit generalization to unheard sounds. \\ 
Lastly, we propose a new architecture for both tasks that allows the agent to fuse the information from spatial maps and audio observations. Compared to the current state-of-the-art models \cite{chen2020soundspaces, chen2020learning}, our model is capable of generalizing better to unheard sounds in both clean and complex audio scenarios setups.
\par
We demonstrate our results on Habitat simulator \cite{habitat19iccv} and its audio-compatible SoundSpaces simulator \cite{chen2020soundspaces} which provides the ability to generate binaural sound signals for the realistic 3D environments of the Replica \cite{straub2019replica} and Matterport3D \cite{chang2017matterport3d} datasets. Our approach improves the state-of-the-art on the unheard static AudioGoal benchmark task~\cite{chen2020soundspaces} with 58\% and 39\%  better success weighted by path length (SPL) and 53\% and 29\% better success rate (SR) on Replica and Matterport3D, respectively.
\par 
Our work won the \href{https://soundspaces.org/challenge}{SoundSpaces Challenge at The Conference on Computer Vision and Pattern Recognition (CVPR) 2021}. We also submitted our main contributions to a Computer Vision Conference under the title "Catch Me If You Hear Me: Dynamical Audio-Visual Navigation in Unmapped Complex 3D Environments with Moving Sounds" \cite{younes2021catch}. 

%% file: chapters/2-background.tex
\chapter{Background}\
\label{chap:background}
\numberwithin{equation}{chapter}
In this Background chapter, we explain the empirical foundations we built on in this thesis, starting with Habitat, the simulator that we used to train an embodied agent on the task we tackle and following with the building blocks of our approach, Reinforcement Learning, Proximal Policy Optimization, Convolutional Neural Network, Gated Recurrent Unit, and Spectrogram Augmentation.
\section{Habitat}
\label{sec:habitat}
Towards Artificial General Intelligence, AI agents need to learn from an egocentric perception like humans, which requires a shift from the 'Internet AI' era where we train the models on static datasets from the internet, to 'Embodied AI' era to enable these AI agents to learn through interaction with environments. Moreover, since training in the real world is costly in terms of time, money, and safety, highly photorealistic simulators emerge to be a reasonable solution to train agents of tomorrow on them before deploying them in the real world. 
\par
\begin{figure}[htpb!]
  \centering
    \includegraphics[width=0.9\textwidth]{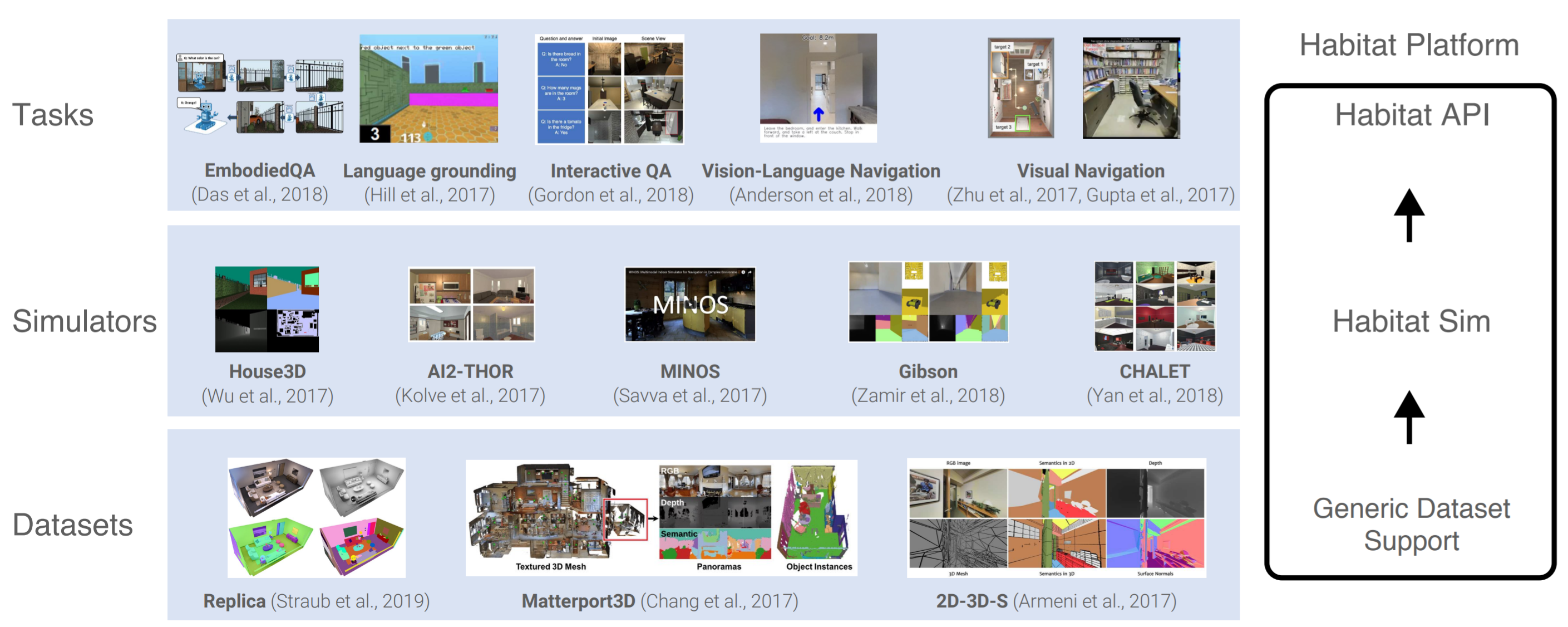}
    \caption{Embodied agents' training stack (right) includes 1. 3D datasets to train the agents on. 2. Simulators that render these datasets and allow to simulate the embodied agent's behavior inside it. 3. Tasks that evaluate the scientific progress in this field. Habitat \cite{habitat19iccv} (left) provides a unified embodied agent platform that includes support of various datasets, a simulator with high performance, and a flexible API making it easier to define and evaluate new tasks \cite{habitat19iccv}.}
    \label{fig:habitat}
\end{figure}

Habitat is a platform for Embodied AI research that enables the training of virtual robots in efficient photorealistic 3D simulations \cite{habitat19iccv}. Habitat platform consists of two essential components: Habitat-Sim and Habitat-Lab. Habitat-Sim is a fast, flexible, efficient, high-performance 3D simulator, which provides generic handling of 3D datasets, sensors, and configurable AI agents. Habitat-sim has the ability to reach 10,000 rendering scene frames per second (fps) multi processes using one GPU only. Habitat-Lab (previously Habitat-API) is a high-level library that aims to ease the development and the benchmarking of end-to-end learning techniques to train the embodied agents on several tasks. Habitat's stack is illustrated in Figure \ref{fig:habitat}.
\par
Tasks like: PointGoal Navigation, ObjectGoal Navigation, Vision-Language Navigation, Active Visual Tracking, Visual Exploration, Embodied Question Answering, and other exciting tasks can be learned by the AI agents using Habitat only. The embodied AI agents need both Habitat \cite{habitat19iccv} and SoundSpaces \cite{chen2020soundspaces} Section (\ref{soundspaces}) platforms to learn the AudioGoal task in which the agent tries to navigate towards a sound-emitting target source using only audio and visual observations.

Despite the presence of several other simulators like AI2-THOR \cite{kolve2017ai2}, iGibson \cite{xia2020interactive}, DeepMind Lab \cite{beattie2016deepmind}, and ThreeDWorld \cite{gan2020threedworld}. Habitat \cite{habitat19iccv} and iGibson are the only ones that offer realism by constructing real-world-based scene instead of game-based scene construction \cite{duan2021survey} with the support of real-world datasets like Gibson V1 \cite{xia2018gibson}, Matterport3D \cite{chang2017matterport3d}, and Replica \cite{straub2019replica}. Nevertheless, Habitat offers 10x more speed than iGibson \cite{duan2021survey}, and with the SoundSpaces platform \cite{chen2020soundspaces} Section (\ref{soundspaces}), Habitat is the most suitable simulator to train embodied agents on the AudioGoal task in highly realistic, complex 3D environments.


\section{Reinforcement Learning}
This section is based on the Reinforcement Learning book \cite{sutton2018reinforcement} and intended to be a brief explanation of Reinforcement Learning; please consider reading the book if you need more details.\par
To enable an AI agent to learn a goal-oriented specific task such as audio-visual navigation in habitat simulator Section \ref{sec:habitat}, the agent must learn by interacting with the surrounding environment until it learns how to perform this task correctly. This kind of learning, where the agent relies on a trial-and-error search and the received delayed reward, is called reinforcement learning.
\par
Reinforcement learning tackles the whole problem of an intelligent agent interacting with an unknown environment to perform a specific task. 
In our audio-visual navigation problem, reinforcement learning provides an end-to-end goal-directed learning technique that induces the agent to learn all the required sub-tasks to solve the whole problem of navigating towards a sound-emitting source like recognizing traversable paths, obstacle avoidance, environment exploration, building geometrical and acoustic maps to memorize previous seen/heard observations, planning the shortest route to the goal, and deciding a set of actions to follow that route.

\subsection{Reinforcement Learning Elements}
There are other components of the Reinforcement Learning system other than the agent and the environment, which are a reward, policy, value function, and model. \\
A \textit{reward R} is a scalar feedback signal the agent receives every step to inform it how well or bad its action was in this step. The agent aims to maximize the \textit{return}, which is the total cumulative sum of the received reward in each running episode. The discounted \textit{return G} can be calculated using the following Equation \ref{eq:return}, where the discounted factor $\gamma \in [0,1]$ is added to avoid infinite returns.
\begin{equation}
    G_t = R_{t+1} + \gamma R_{t+2} + \gamma^2 R_{t+3} + .. = \sum_{k=0}^{\infty} \gamma^k R_{t+k+1} = R_{t+1} + \gamma G_{t+1} 
    \label{eq:return}
\end{equation}
A \textit{policy $\pi$} describes how the agent behaves in the environment. It maps the perceived state $s$ of the environment to action $a$ for the agent to take. The policy can be deterministic or stochastic, specifying a probability for each action for the agent to take for the perceived environment's state. \\
\begin{equation}
    deterministic: \pi(s) =a, stochastic: \pi(a|s) = Pr[A_t = a|S_t = s]
    \label{eq:policy}
\end{equation}
A value function describes how good is it to be in that state by providing an expected accumulative reward the agent can receive starting from this state and following the policy \textit{$\pi$}. There is also an  \textit{action value function $q_\pi$(s,a)} that defines the expected reward the agent can gain if it starts at state \textit{s}, selects action \textit{a}, and follows policy \textit{$\pi$}.
\begin{equation}
    v_\pi(s) = \mathop{{}\mathbb{E}}_\pi[G_t|S_t = s] = \mathop{{}\mathbb{E}}_\pi\left[\sum_{k=0}^{\infty} \gamma^k R_{t+k+1}|S_t = s\right]
    \label{eq:value}
\end{equation}
\begin{equation}
    q_\pi(s,a) = \mathop{{}\mathbb{E}}_\pi[G_t|S_t = s, A_t = a] = \mathop{{}\mathbb{E}}_\pi\left[\sum_{k=0}^{\infty} \gamma^k R_{t+k+1}|S_t = s, A_t = a\right]
    \label{eq:qvalue}
\end{equation}
Thus the connection between \textit{value function $v_\pi$(s)} and \textit{action value function $q_\pi$(s,a)} can be presented by the following equation:
\begin{equation}
     v_\pi(s) = \mathop{{}\mathbb{E}}_\pi[q_\pi(s,\pi(s))]
    \label{eq:connection}
\end{equation}
Finally, there is an optional component of the reinforcement learning system, which is the \textit{model} of the environment. The \textit{model} can be described as a graph to define the transition between the environment's states. Given the state and the action, the \textit{model} predicts the next state and the next reward.

\subsection{Markov Decision Processes}

\begin{figure}[htpb!]
  \centering
    \includegraphics[width=0.9\textwidth,height=0.9\textheight, keepaspectratio]{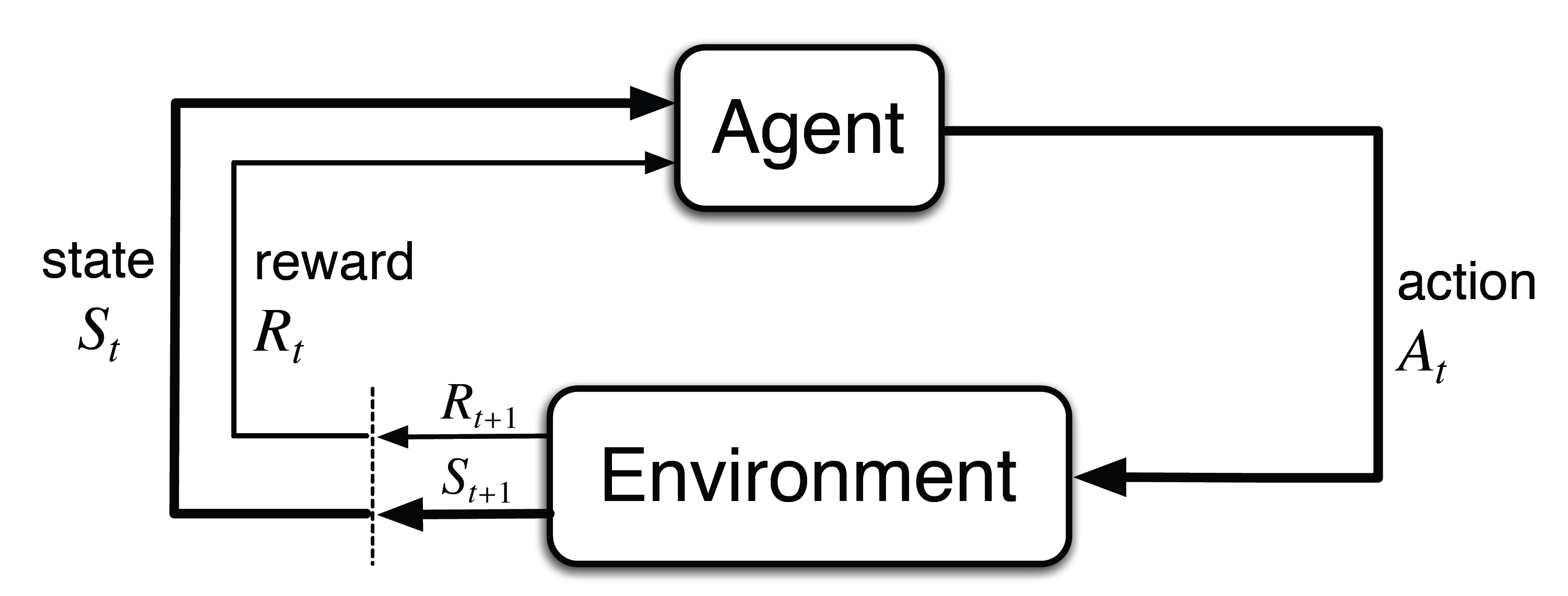}
    \caption{The interaction between agent and environment in Markov Decision Process \cite{sutton2018reinforcement}.}
    \label{fig:rl_system}
\end{figure}
Reinforcement learning approaches tackle the problems described in Markov Decision Processes (MDP) form \cite{bellman1957markovian}. A finite Markov Decision Processes (MDP) form consists of \cite{bellman1957markovian} $\langle \textit{S}, \textit{A}, \textit{p}, \textit{R} \rangle$, where \textit{S} defines the finite number of states, \textit{A} represents the finite number of available actions, \textit{p} is the transition probability function and \textit{R} defines the finite set of rewards. The next state $S_{t+1}$ and the next reward $R_{t+1}$ have discrete probability distributions based only on the current state and action as we have a finite number of states \textit{S}, actions \textit{A}, and rewards \textit{R}. This satisfies the Markov Property in Equation \ref{eq:markov}, which states that the state must include information about all aspects
of the past agent–environment interaction that makes a difference for the future \cite{sutton2018reinforcement} or in other words, the next state and reward is independent of the past states and actions, given the current state and action. Agent-environment interaction in Markov Decision Process is illustrated in Figure \ref{fig:rl_system}.
\begin{equation}
    Pr\{S_{t+1}, R_{t+1}|S_{t}, A_{t}\} = Pr\{S_{t+1}, R_{t+1}|S_{t}, A_{t}, ...., S_{0}, A_{0}\}
    \label{eq:markov}
\end{equation}

Reinforcement learning agents aim to maximize the received numerical reward signal by learning which actions to take in each situation yield the most reward.
The problem of audio-visual navigation covers the three main aspects of the simplest Markov Decision Processes form \cite{bellman1957markovian}: 1. Sensation: The agent can sense the environment by receiving visual and audio observations. 2. Action: The agent can take actions from Habitat's available set of actions that affect the environment and its following observed states. 3. Goal: The agent has a goal to navigate towards the sound-emitting source. According to the authors of \cite{sutton2018reinforcement}, any method that can solve this kind of problem is considered a reinforcement learning method.

\par

\subsection{Exploration Exploitation Dilemma}

One of the main challenges that any reinforcement learning agent face is it has to \textit{exploit} the actions that it has previously tried and led to a positive reward, but at the same time, it has to \textit{explore} new actions to get possible higher rewards in the future. To balance the trade-off between \textit{exploration} and \textit{exploitation}, the agent must \textit{explore} more at the beginning of the training, and then it has to gradually decrease the \textit{exploration} as the agent becomes more particular about the expected reward of the set of actions. $\epsilon-greedy$ method is one of the approaches that performs this idea with a coupled decay schedule for $\epsilon$. The agent selects the greedy action with a probability $(1 - \epsilon)$, and with a probability $\epsilon$, the agent takes a random action. The value of $\epsilon$ decays with time to decrease exploration when the agent's estimate about the value of taking specific action in a specific state becomes more accurate; an example of this decay schedule can be found in Equation~\ref{eq:schedule}, where \textit{t} is the current round while \textit{T} is the number of the maximum rounds. \textit{Exploration} at the beginning works on enriching the agent's knowledge about the values of the available set of actions, while \textit{exploiting} is favorable later in the training as it is the way to maximize the received reward. This technique is not preferable in non-stationary problems as exploration is needed even later in training as the setup changes with time. 
\begin{equation}
    \epsilon_{t+1} = (1- \frac{t}{T})\epsilon_{init}
    \label{eq:schedule}
\end{equation}

\subsection{Bellman Optimality Equation}

The Bellman equation \cite{bellman1958dynamic} provides the relationship between the value of a given state and the values of its successor states in a Markov Decision Processes problem.
Bellman equation for the \textit{value function $v_\pi$(s)} is defined as:
\begin{equation}
    v_\pi(s) = \sum_{a} \pi(s|a) \sum_{s^{'},r}p(s^{'},r|s,a)[r+\gamma v_\pi(s^{'})]
    \label{eq:valueBellman}
\end{equation}
Similarly, the Bellman equation for the \textit{action value function $q_\pi$(s,a)} is defined as:
\begin{equation}
    q_\pi(s,a) = \sum_{s^{'},r} p(s^{'},r|s,a) \left [r + \gamma \sum_{a^{'}}\pi(a^{'}|s^{'})q_\pi (s^{'},a^{'}) \right ]  
    \label{eq:actionBellman}
\end{equation}

An optimal \textit{policy $\pi_{*}$} which achieves the most reward over the long run is considered to be the solution of a reinforcement learning problem. This \textit{policy $\pi_{*}$} is called optimal if and only if its expected return for all states is better than or equal the expected return of any other \textit{policy $\pi$} or in other words for all $s \in S$, \textit{value function $v_{\pi*}$(s)} has to be greater than or equal \textit{value function $v_{\pi}$(s)} for all $\pi$. A search among all policies is needed to satisfy this condition, but this is impractical as the number of policies can be infinite. But the Bellman optimality equation \cite{bellman1958dynamic} states that the value of a state $v_{*}(s)$ under an optimal \textit{policy $\pi_{*}$} must equal the expected return for the best action $a$ from that state $s$ and following the optimal \textit{policy $\pi_{*}$} \cite{sutton2018reinforcement}. There is a unique solution for the Bellman optimality Equation \ref{eq:valueBellmanopt} for $v_{*}(s)$ in case of finite MDP. This means it is easier to determine the optimal \textit{policy $\pi_{*}$} once we have the optimal \textit{value function $v_{*}(s)$}  because, for each state $s$, there will be at least one action that obtains the maximum in the Bellman optimality equation. 
Bellman optimality equation for the \textit{value function $v_*$(s)} is defined as:
\begin{equation}
    v_*(s) = \max_a \sum_{s^{'},r} p(s^{'},r|s,a)[r + \gamma v_*(s^{'})]
    \label{eq:valueBellmanopt}
\end{equation}
Similarly, the Bellman optimality equation for the \textit{action value function $q_*$(s,a)} is defined as:
\begin{equation}
    q_*(s,a) = \sum_{s^{'},r} p(s^{'},r|s,a) [r + \gamma \max_{a^{'}}q_* (s^{'},a^{'}) ]  
    \label{eq:actionBellmanopt}
\end{equation}
The Bellman optimality equation for the \textit{action value function} $q_*$(s,a) in Equation \ref{eq:actionBellmanopt} makes selecting the optimal actions easier as the agent can choose the action that maximizes $q_*$(s,a). Most reinforcement learning approaches can be considered as approximates to solve the Bellman optimality equations.

\subsection{Actor-Critic Methods}
There are three main classes of solving reinforcement learning problems, model learning, value function learning, and policy learning. The model learning approach can be more straightforward or more challenging than the other two methods based on the task itself, but it is hard to extract an optimal policy using it. The action-value methods learn the values of the actions then depend on these values for action selection. The third method is policy learning, which directly learns a parameterized policy that allows action selection without action-value estimates. The parameterized policy gradient methods with parameters $\theta$ can be modeled as:
\begin{equation}
    \pi(a|s,\theta) = Pr[A_t = a|S_t = s, \theta_t = \theta]
    \label{eq:policyG}
\end{equation}
The parameters $\theta$ are updated based on the gradient ascent of some scalar performance measure $J(\theta)$ as follows:
\begin{equation}
    \theta_{t+1} = \theta_t + \alpha  \widehat{\nabla J(\theta_t)} 
    \label{eq:thetaUpdate}
\end{equation}
These policy learning methods have some advantages over the action-value learning methods as they are capable of learning probabilities for taking the actions, handling large and continuous action spaces, and asymptotically reaching deterministic policies. Parameterized policy methods are also supported with the Policy Gradient theorem, which provides a formula that describes the effect of policy parameter that does not have derivatives of the distribution of the state on the performance. 
The policy gradient theorem for the episodic case states that:
\begin{equation}
    \nabla J(\theta) \propto \sum_s \mu(s)\sum_a q_\pi \nabla \pi(a|s,\theta)
    \label{eq:theorem}
\end{equation}
where the distribution $\mu$ is the on-policy distribution under $\pi$.\\
A group of sub-class methods of Policy Gradient approaches is called Actor-Critic methods; this group of methods aims to reduce the high variance in the vanilla policy gradient methods like REINFORCE \cite{williams1992simple} as it is unbiased and learns slowly by using a bootstrapping, which updates the estimated value of the state depends on the estimated values of the subsequent states. The actor-critic methods learn both policy and value function where the actor is responsible for learning the policy, and the critic learns the value function with bootstrapping. The critic state-value bootstrapping introduces bias, but it is helpful to decrease the variance and speed up the training.

\subsection{Proximal Policy Optimization}
\label{sec:ppo}
The primary optimization objective function which policy gradient methods usually use to learn is defined as:
\begin{equation}
    L^{PG}(\theta) = \mathop{{}\hat{\mathbb{E}}_t} \left [ \log \pi_\theta (a_t|s_t) \hat{A}_t \right]
    \label{eq:objective}
\end{equation}
Where the expectation $\mathop{{}\hat{\mathbb{E}}_t}$ indicates the empirical average over a finite batch of samples, $\hat{A}_t$ is the advantage function which provides an estimate comparing the actual return the agent got with the expected return from the network. Multiplying the log probabilities of the policy $\pi_\theta$ output with the advantage function provides the optimization function. If the advantage function is positive, which means the action performed better than expected, then the policy gradient would be positive to increase the probability of choosing this action again once the agent arrives at a similar state. While if the advantage is negative, that means the action performed worse than average, then the policy gradient would be negative, which will decrease the probability of selecting this action if the agent encounters a similar state in the future.  However, the primary problem with this objective function is that it leads to destructive large policy updates, which causes instability in training.\\
The authors of Trust Region Policy Optimization (TRPO) \cite{schulman2015trust} proposed a new surrogate objective function with a constraint for the size of policy's update to maximize to solve the problem of destructive extensive policy updates that face the primary policy gradient objective function in Equation \ref{eq:objective}. The TRPO surrogate optimization function and its constraint are defined as:
\begin{equation}
    \max_\theta \mathop{{}\hat{\mathbb{E}}_t} \left [ \frac{\pi_\theta (a_t|s_t)}{\pi_{\theta_{old}} (a_t|s_t)} \hat{A}_t \right]
    \label{eq:trpo}
\end{equation}
\begin{equation}
    subject \: to \; \mathop{{}\hat{\mathbb{E}}_t}  [KL[\pi_{\theta_{old}}(.|s_t),\pi_{\theta}(.|s_t)] ] \leq \delta
    \label{eq:constraint}
\end{equation}
The Kullback-Leibler (KL) divergence \cite{kullback1951information} constraint limits the update from taking large steps away from the old policy, but this hard KL constraint usually adds overhead to optimization and training. 
\par
\begin{figure}[htpb!]
  \centering
    \includegraphics[width=0.9\textwidth,height=0.9\textheight, keepaspectratio]{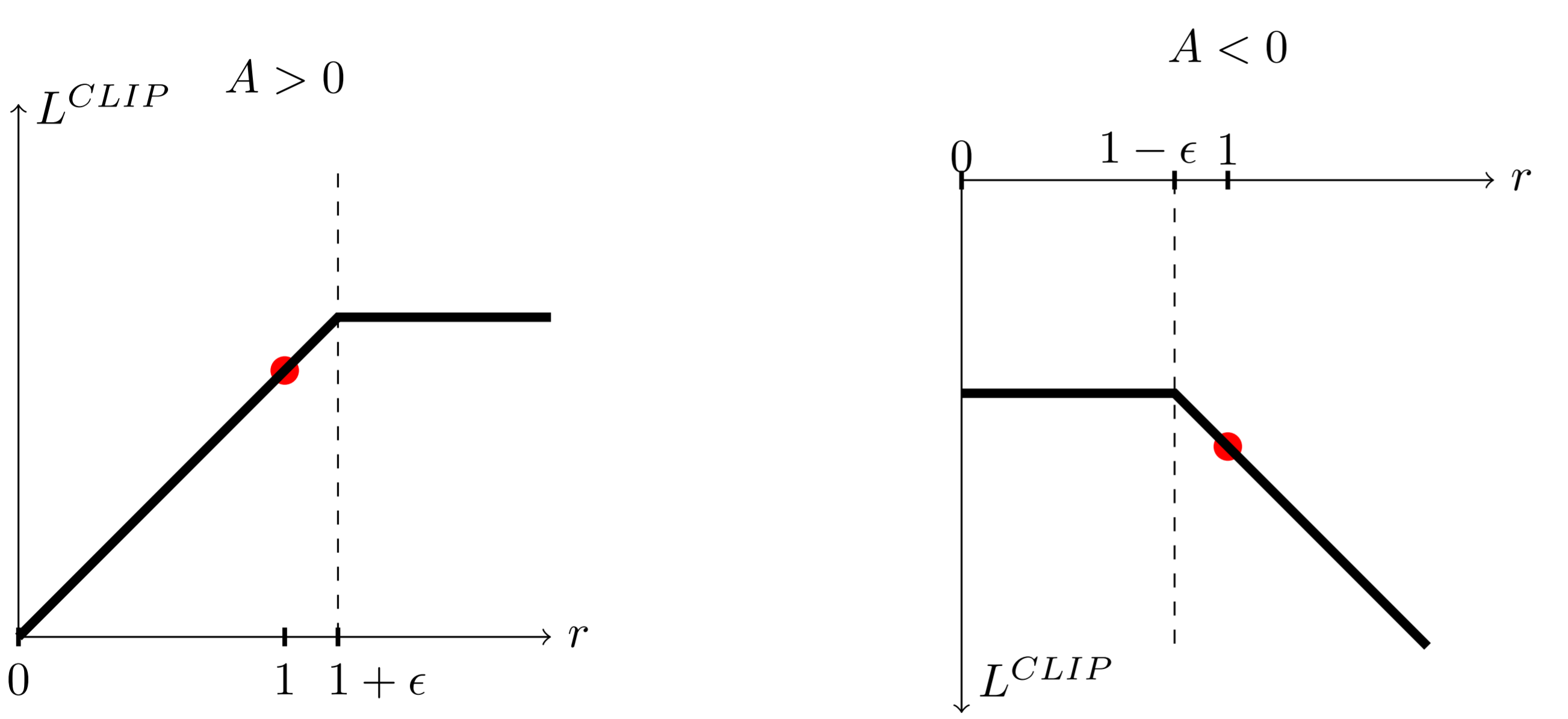}
    \caption{On the left side, the plot shows how the clipped loss function behaves when the advantage function $A_t$ is greater than zero, once the ratio $r_t$ between the new and old policy is greater than $1 + \epsilon$, the function clips the loss. While on the right side, the function clips the loss if the advantage function $A_t$ is lower than zero and the ratio $r_t$ is lower than $1 - \epsilon$ \cite{schulman2017proximal}.}
    \label{fig:clipped}
\end{figure}

Proximal Policy Optimization (PPO) \cite{schulman2017proximal} builds on the work from TRPO by trying to integrate the constraint directly into the objective function. The primary objective function of PPO is:
\begin{equation}
   L^{CLIP}(\theta) = \mathop{{}\hat{\mathbb{E}}_t}[\min(r_t(\theta)\hat{A}_t ,clip(r_t(\theta), 1 - \epsilon, 1 + \epsilon)\hat{A}_t)]
    \label{eq:clip}
\end{equation}
\begin{equation}
    where \: r_t(\theta) = \frac{\pi_\theta (a_t|s_t)}{\pi_{\theta_{old}}(a_t|s_t)}
    \label{eq:rt}
\end{equation}
the $r_t(\theta)$ denotes the probability ratio between new and old policy updates, the clip operator constraint the moving large steps from the current policy by forcing the ratio to be in the range of $[1 - \epsilon, 1 + \epsilon]$, where $\epsilon$ usually equals $0.2$. This means the probability ratio $r_t(\theta)$ between new and old policy has to be in the small range between $[0.8, 1.2]$ to avoid large destructive policy updates.  The $\min$ function plays a crucial rule in case the advantage function is negative and the probability ratio $r_t(\theta)$ is positive; in this case, the action did worse than the average hence the negative advantage function, but since the $r_t(\theta)$ is positive the new policy increases the likelihood to take this action in similar states. In this case, the $r_t(\theta)\hat{A}_t$ is lower than the clipped part; hence the $\min$ function will select it, and this will undo the last gradient step by moving with the same amount in the negative direction. An illustration of this clipped function can be shown in Figure \ref{fig:clipped}, while the full PPO objective function that used in training is defined as:
\begin{equation}
   L^{CLIP+VF+S}(\theta) = \mathop{{}\hat{\mathbb{E}}_t}[L^{CLIP}(\theta) - c_1 L_t^{VF}(\theta) + c_2 S[\pi_\theta](s_t)]
    \label{eq:full_obj}
\end{equation}
The final Equation \ref{eq:full_obj} consists of two more terms other than the clipped term presented in Equation \ref{eq:clip} which are $L_t^{VF}(\theta)$ that presents the squared error loss function  of the value functions and the second term $S[\pi_\theta](s_t)$ acts as an entropy bonus to encourage training, $c_1$ and $c_2$ are the weights for these terms in the Equation \ref{eq:full_obj}. The complete PPO algorithm \cite{schulman2017proximal} can be found at Algorithm \ref{alg:ppo}
\input{tables/ppo}

\section{Convolutional Neural Network}
\label{sec:cnn}
Classical Machine Learning algorithms require manual feature extraction. That step has slowed down the progress in Machine Learning as it is hard to do and requires expert knowledge in the input data field. Convolutional Neural Network (CNN) is a unique network architecture for Deep Learning which eliminates the need for manual feature extraction by learning directly from the data. The Convolution layer or the Kernel or the Filter is the main component of CNN; that kernel (filter) is a matrix with usually a smaller dimensional than the input data. The role of this kernel is to extract the essential features of the input data by sliding over all of its dimensions and performing dot product between the kernel's values and the corresponding slid part in the input data point. Each layer's convoluted feature map output serves as an input to the next layer in the network. The first layers tend to learn low-level features, while the later layers learn more advanced high-level features. The network learns the values of those filters (also known as weights) using the backpropagation algorithm \cite{leung1991complex} during the training process.
\par
The convolution operation is defined as:
\begin{equation}
    S(i, j) = (K * I)(i, j) = \sum_m \sum_n I(i - m, j - n)K(m, n) 
    \label{1}
\end{equation}
It is usually implemented by flipping the kernel and using the following Cross-correlation's equation:

\begin{equation}
    S(i, j) = (I * K)(i, j) = \sum_m \sum_n I(i + m, j + n)K(m, n) 
    \label{2}
\end{equation}
where $I$ in Equation \ref{1} and Equation \ref{2} represents the input data point (e.g., image), $i$ and $j$ represent the indices of a point on the image, but $m$ and $n$ represent the indices of a point on the kernel,  while $K$ represents the kernel that slides over the whole input data calculating the dot product between its weights and the corresponding values of the data point. Then it sums the output of that dot product operations to a single value in the $feature$ $map$. The following Figure \ref{fig:cnn} illustrates more about this convolution process.
\begin{figure}[htpb!]
  \centering
    \includegraphics[width=0.9\textwidth,height=0.9\textheight, keepaspectratio]{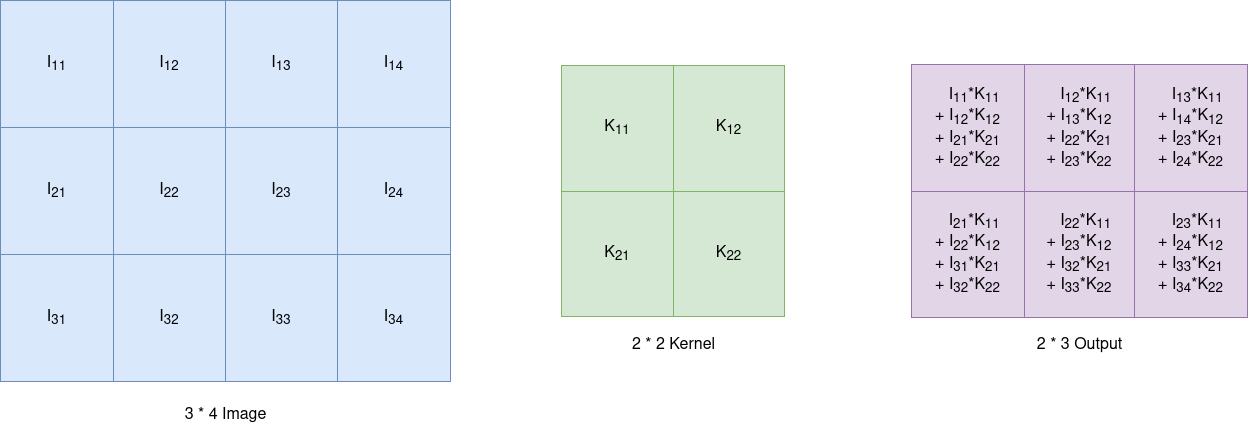}
    \caption{An illustration of the cross-correlation, where the left matrix $I$ is the input image with size $(3,4)$, the middle $K$ is the kernel with size $(2,2)$ and the output matrix is on the right with size (2,3).}
    \label{fig:cnn}
\end{figure}
\section{Gated Recurrent Unit}
\label{sec:gru}
Convolutional Neural Network (CNN), as explained in Section \ref{sec:cnn} is a feed-forward architecture which means it does not allow cycles in its connectivity graph, i.e., the data transfers from one hidden layer to the next one in the direction of the output only. This is useful when dealing with non-sequential input data, and the previous inputs or features are irrelevant to the current time step output. However, in our problem, the previous environment's states and observations might help our agent perform better. This can be seen as a memory of our agent that helps it keep the valuable information or features from previous observations and consider them while selecting the following action. Recurrent Neural Network (RNN) \cite{rumelhart1985learning, rumelhart1986learning} allows cycles in its connectivity graph so, it is suitable for dealing with sequence data. Nevertheless, RNN architecture suffers from vanishing and exploding gradients problem depending on the values of the eigenvalues of the Weight matrix; if the eigenvalues are smaller than one, this could lead to vanishing gradient problem, while if the eigenvalues are greater than one, this could lead to exploding gradient problem.  One possible solution to the exploding gradient problem is clipping the gradients above a specific threshold.\par
\begin{figure}[htpb!]
  \centering
    \includegraphics[width=0.9\textwidth,height=0.9\textheight, keepaspectratio]{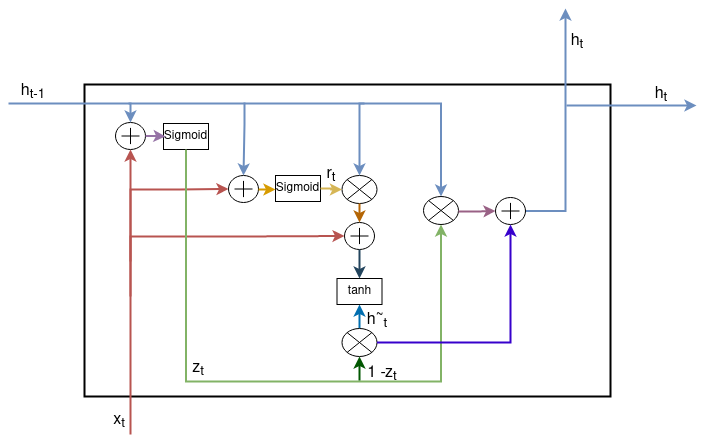}
    \caption{Gated Recurrent Unit (GRU) architecture for a single hidden state.}
    \label{fig:gru}
\end{figure}

Gated Recurrent Unit (GRU) \cite{cho2014learning} provides an architecture that solves the vanishing gradient problem. GRU achieves that using two gates, an update gate and a reset gate; these two gates allow the network to remember or forget the previous information adaptively. Let us consider a case with a single hidden layer of GRU network like the one in Figure \ref{fig:gru}, the reset gate for step $t$ is defined as: 
\begin{equation}
    r_t = \sigma( W_r x + U_r h_{t-1})
    \label{eq:reset}
\end{equation}
where $r_t$ is the reset gate for the current step $t$, $\sigma$ is  a logistic sigmoid activation function, $W_r$ is the learned weight matrix for input $x$ while $U_r$ is the learned weight matrix for the previous step hidden unit output $h_{t-1}$. This gate decides the amount of previous information for the network to forget. The update gate is calculated similarly but with different weights $W_z$ and $U_z$ as presented in the following equation:
\begin{equation}
    z_t = \sigma( W_z x + U_z h_{t-1})
    \label{eq:update}
\end{equation}
The current information this hidden unit has can be presented as:
\begin{equation}
    \tilde{h_t} = \tanh( W x + U[r_t \odot h_{t-1}])
    \label{eq:hidden}
\end{equation}
where $\tanh$ is an activation function, $\odot$ is element-wise Hadamard product between the current step reset gate $r_t$ calculated in Equation \ref{eq:reset} with the previous time step hidden layer output $h_{t-1}$, then we multiply the output of this product with the weight matrix $U$, we add the output of this product to the product of input $x$ with the weight matrix $W$. Finally, we pass this sum to the $\tanh$ activation function. In this equation, if the reset gate $r_t$ is close to zero, the hidden network will only ignore the previous step's hidden state and rely on the current input data. The update gate $z_t$ controls the amount of information from the previous hidden state that will transfer to the current hidden state using this equation:
\begin{equation}
    h_t = z_t h_{t-1} + (1 - z_t) \tilde{h_t}
    \label{eq:hiddenFinal}
\end{equation}
\section{SpecAugment}
\label{sec:specaugment}
Learning techniques tend to overfit the data by learning non-desirable features that do not help in generalization to other data points. One possible solution to this overfitting problem is data augmentation, where we try to create random deformations in the available data before passing them to the network so the network can focus on learning the valuable features that help in generalization. While data augmentation techniques in image data can be straightforward, such as rotating, flipping, zooming, random cropping, shifting, or brightness changing, this is not the case in audio data as it is hard to come up with a technique that does not affect the critical features of the input. It is also easier in image data than in audio data to judge the quality of augmentation and whether the main features are still included or not. However, the authors of Specaugment \cite{park2019specaugment} presented three techniques to augment data on the spectrogram of the sound data.
\par
\begin{figure}[htpb!]
  \centering
    \includegraphics[width=0.9\textwidth,height=0.9\textheight, keepaspectratio]{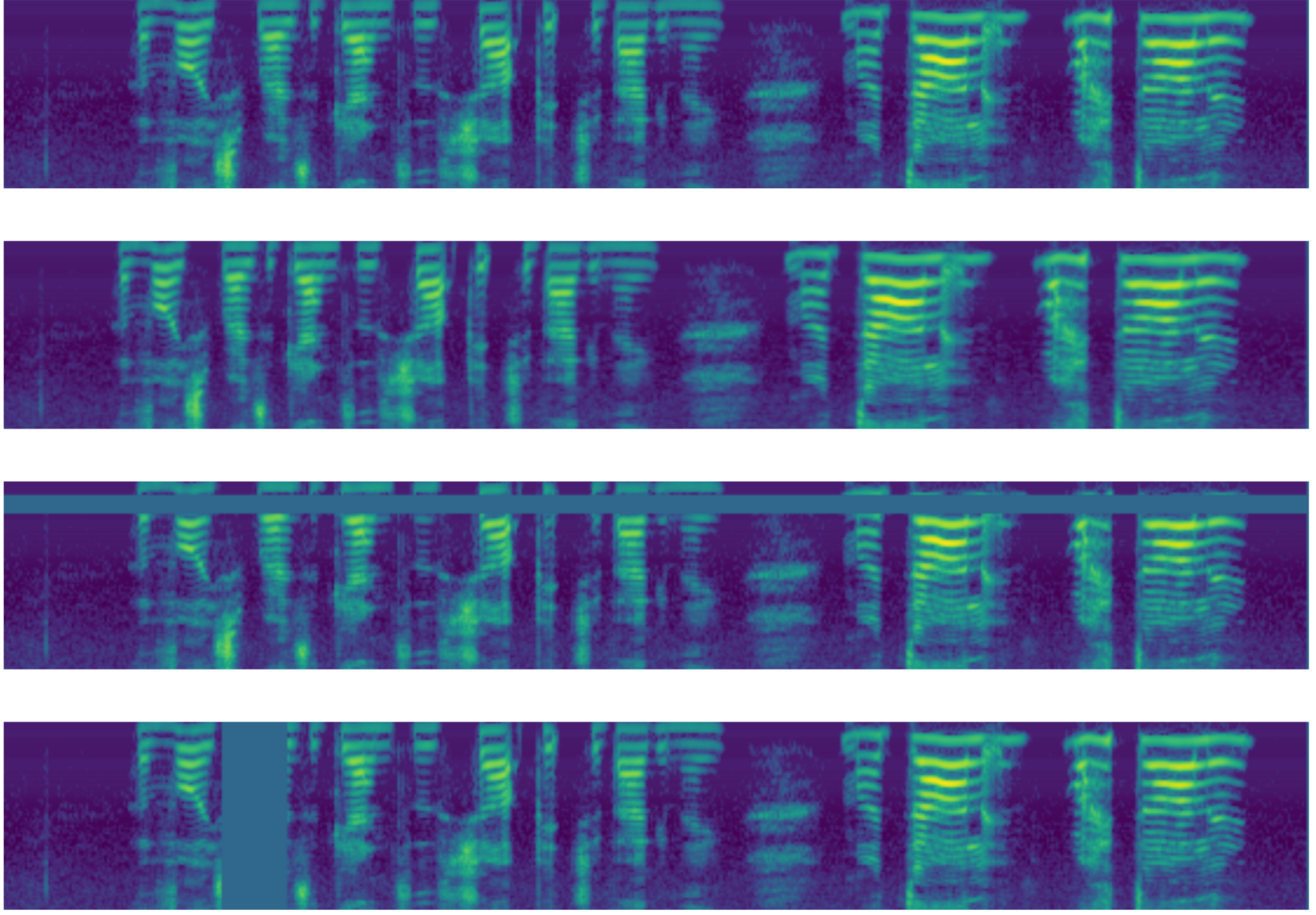}
    \caption{Illustration of the three Specaugment augmentation techniques, top image is the input followed by time wrapping, then frequency masking, and then the bottom one represents the time masking  \cite{park2019specaugment}.}
    \label{fig:specaugment}
\end{figure}

Specaugment \cite{park2019specaugment} introduced Time Wrapping, Frequency Masking, and Time Masking as feature augmentation techniques for sound data. The sound's spectrogram can be viewed as an image where the horizontal axis represents the time, and the vertical axis represents the frequency. Figure \ref{fig:specaugment} shows the output spectrogram of the three methods.

\begin{itemize}
    \item Time Wrapping: wraps a random point on the horizontal line that passes through the center of the image to the right or the left. The point has to be within the range of $(W, \tau - W)$ and to be wrapped by a distance $w$, where $W$ is the time wrap parameter, $\tau$ is the time steps of the spectrogram, or in other words, the width of the image and  $w$ is a random distance drawn from a normal distribution from 0 to $W$.
    \item Frequency Masking: masks $f$ consecutive frequency channels starting from $f_0$ where $f$ is randomly sampled from a normal distribution between 0 and $F$ the frequency mask parameter while $f_0$ is randomly chosen from [0, $\upsilon$ - $f$) where $\upsilon$ is the number of frequency channels or in other words the height of the image.
    \item Time Masking: like Frequency Masking, it masks $t$ consecutive time steps starting from $t_0$ where $t$ is randomly sampled from a normal distribution between 0 and $T$ the time mask parameter while $t_0$ is randomly chosen from [0, $\tau$ - $t$) where $\tau$ is the number of time channels or in other words the width of the image.

\end{itemize}

%% file: tables/ppo.tex
\begin{algorithm}

\SetAlgoLined
\For{iteration = 1,2,... }
{
    
\For{actor = 1,2,...,N}{
Run policy $\pi_{\theta_{old}}$ in environment for T timesteps \\
Compute advantage estimates $\hat{A}_1$,..., $\hat{A}_T$\\
}
Optimize surrogate $L$ wrt $\theta$, with $K$ epochs and minibatch size $M \leq NT$ \\
$\theta_{old } \leftarrow \theta$\\

}
 \caption{PPO Algorithm \cite{schulman2017proximal}}
 \label{alg:ppo}
\end{algorithm}

%% file: chapters/3-relatedwork.tex
\chapter{Related Work}
\label{chap:relatedwork}
Following the goal of having AI robotic agents that can see, move, speak, hear, and interact with the surrounding environments, recent work in Artificial Intelligence has been shifted from Internet AI, where much work has been done on static datasets (e.g., ImageNet \cite{deng2009imagenet}, COCO \cite{lin2014microsoft}, KITTI \cite{geiger2012we}) to Embodied AI, where the agent interacts with the simulated realistic 3D environments in order to give the agent a learning experience like the infants who learn by interacting with the surroundings. 
\par
Embodied AI aims to incorporate several AI subfields which have been addressed solely before, like Computer Vision, Natural Language Processing, Navigation, Reinforcement Learning, Robotics, and Physics-based simulations. This new era of AI depends on simulators like AI2-THOR \cite{kolve2017ai2}, iGibson \cite{xia2020interactive}, and Habitat \cite{habitat19iccv} and on datasets like Gibson V1 \cite{xia2018gibson}, Matterport3D \cite{chang2017matterport3d}, and Replica \cite{straub2019replica} to speed up the progress in that field.  
\par
The research in embodied AI tackles a variety of tasks like PointGoal Navigation \cite{gupta2017cognitive,chen2019behavioral,wijmans2019dd}, ObjectGoal Navigation \cite{chaplot2020object, yang2018visual, du2021vtnet}, Vision-Language Navigation \cite{anderson2018vision,fried2018speaker,chen2021topological}, Active Visual Tracking \cite{luo2018end,zhong2021towards}, Visual Exploration \cite{pathak2017curiosity,jayaraman2018learning,chaplot2020learning}, Embodied Question Answering \cite{das2018embodied, yu2019multi, wijmans2019embodied}, and other interesting tasks. And 
despite the fact that audio data is a crucial signal for navigation and the use of it provides blind people spatial navigation capability comparable to sighted people \cite{fortin2008wayfinding}, most of the previously mentioned work on embodied AI rely solely on vision-based navigation to perform the required tasks \cite{jain2020cordial,zhu2017target,zhu2017visual} leaving behind an important sensory signal like audio. 
\par
Recent work has tried to use that pivotal signal in unmapped complex 3D  environments by introducing the AudioGoal task, in which the agent is required to navigate using the audio and visual signals to a sound-emitting target source's location \cite{chen2020soundspaces,gan2020look}. More tasks have been introduced following the AudioGoal task like Semantic Audio-Visual Navigation \cite{chen2021semantic}, Active Audio-Visual Source Separation \cite{majumder2021move2hear}, Audio-Visual Floor Plan Reconstruction  \cite{purushwalkam2020audio}, Audio-Visual Dereverberation\cite{chen2021learning}, and Curiosity based exploration via Audio-Visual Association \cite{dean2020see}.
\par
Although some prior work explored using sound in different setups like localization of sound sources in robotics for both static~\cite{nakadai1999sound, nakadai2000active, nakadai2001epipolar, rascon2017} and dynamic sound sources~\cite{grohn2002static, portello2012active, valin2004localization, evers2015bearing}, localization of sound objects in videos \cite{tian2018audio, Arandjelovic_2018_ECCV} and using sound in virtual reality (VR) setup \cite{massiceti2018stereosonic, gunther2004using}. However, none of them has tackled the audio-visual navigation problem in unmapped complex 3D environments. 
\section{Audio-Visual Navigation}
In order to train the embodied agents to navigate towards the sound source in unmapped complex realistic 3D environments, the authors of \cite{chen2020soundspaces} have presented SoundSpaces which is a platform to enable sound in Habitat \cite{habitat19iccv}  simulator by providing binaural room impulse responses for the publicly available 3D photorealistic datasets Replica \cite{straub2019replica} and Matterport3D\cite{chang2017matterport3d}. Those binaural room impulse responses allow the insertion of any arbitrary sound inside any location on the grid and provide the agent with the ability to hear the sound from its current location by convolving the goal sound with the binaural room impulse response between the location of the target sound source and the agent's current location. \par 

The work of \cite{chen2020soundspaces} has also shown that audio helps in navigation by comparing the results of their model and other baselines on the two tasks PointGoal and AudioPointGoal on both Replica \cite{straub2019replica} and Matterport3D \cite{chang2017matterport3d} datasets. Using audio in the AudioPointGoal task has helped the agent to achieve better Success weighted by Path Length (SPL) \cite{anderson2018evaluation}. Finally,  the authors of \cite{chen2020soundspaces} have proposed a multi-modal reinforcement learning approach to train the agent to navigate towards the sound emitting source using only audio and visual observations.  \par

While the approach of \cite{chen2020soundspaces} acts step-wise in each iteration, the authors of \cite{chen2020learning} introduced waypoints that are dynamically set and learned during the training of the navigation's policy. This work \cite{chen2020learning} has also introduced an auditory memory which provides the agent with a record of what it has heard while navigating inside the environment. These two novel ideas, as long as a geometric map, improved the results on the AudioGoal task in both heard and unheard setups and became the new state-of-the-art for audio-visual navigation. We built our work on this audio-visual navigation task and expanded it to include more challenging scenarios and moving sound sources.

\section{Semantic Audio-Visual Navigation}
The work of \cite{chen2020soundspaces, chen2020learning,gan2020look} assumes that the sound source will be constantly played throughout the episode. This setup can be beneficial in some cases, like when the agent is required to navigate towards the fire alarm once it goes off, but there are other cases when the agent needs to go to the dishwasher or the microwave after hearing their alarms for a short period of time. This type of case requires another setup where the goal audio is not constantly played, and there is a semantic meaning between the audio and the visual appearance of the audio-emitting source.  The work of \cite{chen2021semantic} has provided this setup for that kind of cases by expanding the SoundSpaces simulator to include semantically grounded audios for 21 objects in Matterport3D dataset \cite{chang2017matterport3d} and playing the acoustic events occasionally. While the semantic audio-visual navigation task aims to tackle the short period occasionally auditory events of static sound sources, our work focuses more on covering the noisy, complex audio scenarios and dynamic sound sources. 

\section{Active Audio-Visual Source Separation}
Both audio-visual navigation \cite{chen2020soundspaces, chen2020learning,gan2020look} and semantic audio-visual navigation \cite{chen2021semantic} tasks assume only one sound is played during the episode, and the agent is required to navigate to the sound source's location. However, in active audio-visual source separation \cite{majumder2021move2hear} task, there is more than one audio source of different types (e.g., background noise, speech, music) play simultaneously, and the agent needs to move in an intelligent way to separate the input target monaural sound source (e.g., instrument or a specified human speaker) from the other distractors sounds within a fixed time limit. This setup is different from our approach to tackle the complex audio scenarios, as, in this setup, the agent gets the target sound source as input, but in our setup, the agent does not receive this information and has to learn the constantly sound-emitting goal target independently and to navigate towards it despite the noisy observations or the distractors sounds.  

\section{Audio-Visual Floor Plan Reconstruction }
In the floor plan reconstruction task, the authors of \cite{purushwalkam2020audio} use the audio as an extra signal to visual input to enrich the perception about the surrounding environment. Audio can reveal the existence of a kitchen if there is a sound of a mixer or the existence of a bathroom if the agent hears a sound of a toilet flush. This ability can accelerate the floor plan reconstruction by a large margin compared to methods that depend on visual perception only. Their audio-visual map model can reconstruct the floor plan with 66 \% accuracy after viewing a few glimpses spanning 26\% of an area of the test split of Matterport3D dataset \cite{chang2017matterport3d}. In contrast to the floor plan reconstruction, we aim to use audio and visual observations to navigate towards the sound-emitting source instead of just using these observations to reconstruct a floor plan.

\section{Curiosity-based Exploration via Audio-Visual Association}
Most of the embodied AI agents use reinforcement learning algorithms to learn the tasks, and exploration is one of the challenges that face those algorithms to learn an optimal policy. Consequently, the work of \cite{dean2020see} proposed a form of curiosity that rewards the agent for exploring novel associations between audio and vision. This type of intrinsic rewards allows the agent to explore more by exploiting the relationship between the different sensory modalities. The result of that better exploration can be observed on the trained agents as they learned a better policy which allow them to achieve a better extrinsic reward at the end; in addition to that, those agents have more robustness to the noise compared to other agents, which trained using another exploration techniques that rely only on the difference between predicted and real future. 

%% file: chapters/4-datasets.tex
\chapter{Datasets}
\label{chap:dataset}
Training Embodied AI agents in real-world environments is costly in terms of time and money. Hence researchers came up with the idea of sim2real, in which they train their agents on simulators and then deploy them in the real world. One of the most efficient simulators for this purpose is Habitat \cite{habitat19iccv}. Habitat supports the most diverse and photo-realistic indoor 3D environments datasets Matterport3D \cite{chang2017matterport3d}, and Replica \cite{straub2019replica}. SoundSpaces \cite{chen2020soundspaces} supports both datasets, allowing the AI agents to train on audio-visual tasks on Habitat.

\section{Replica}
\begin{figure}[htpb!]
  \centering
    \includegraphics[width=0.9\textwidth]{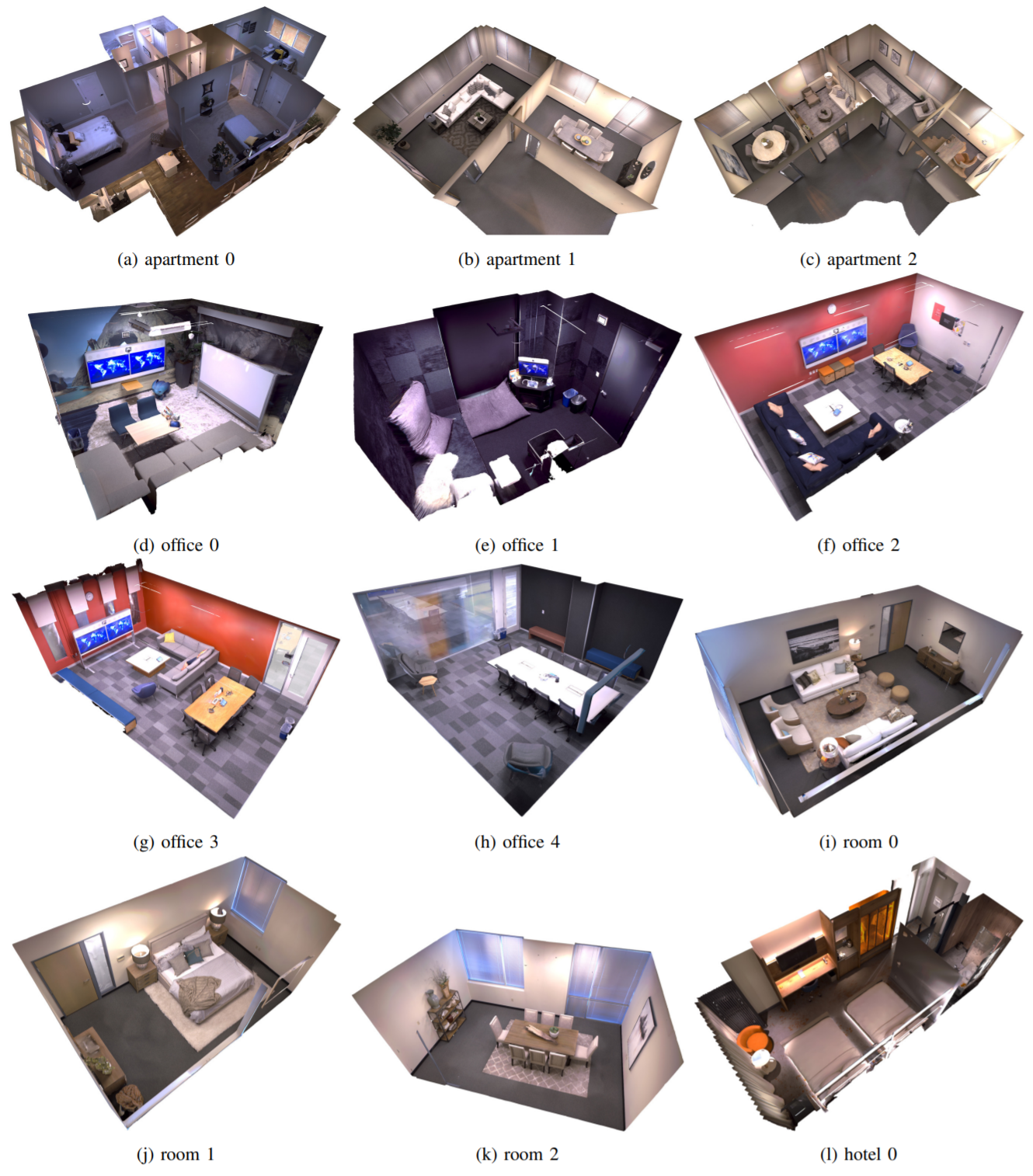}
    \caption{A variety of 12 semantically distinct reconstructions of Replica dataset ~\cite{straub2019replica}.}
    \label{fig:replica1}
\end{figure}

\begin{figure}[htpb!]
  \centering
    \includegraphics[width=0.9\textwidth]{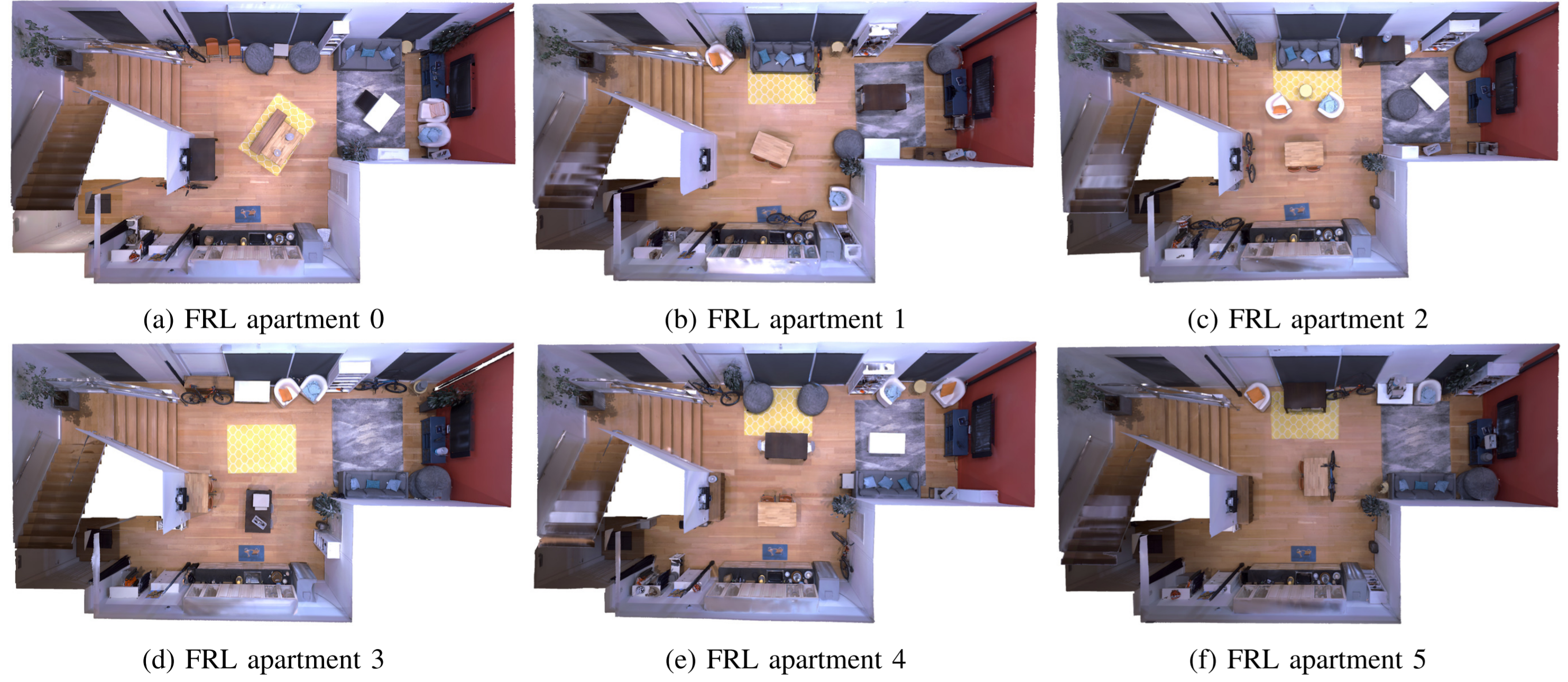}
    \caption{A group of 6 scenes of the FRL apartment from the Replica dataset with the contents rearranged to mimick the exact same scene but at different points in time ~\cite{straub2019replica}.}
    \label{fig:replica2}
\end{figure}

With its 18 highly photorealistic 3D indoor scenes shown in Figure \ref{fig:replica1} and Figure \ref{fig:replica2}, Replica \cite{straub2019replica} serves as a dataset for training embodied agents for geometrically, visually, and semantically realistic tasks. It consists of a hotel room, three rooms of apartments, two multi-room apartment spaces, a 2-floor house, five office rooms, and six different setups of the FRL apartment. These scenes contain high-resolution high-dynamic-range (HDR) textures, a dense mesh, glass reflectors, planar mirrors, per-primitive semantic class, and instance information. Its 35 rooms have $92k$ color resolution $[\frac{pixel}{m^2}]$, $6k$ geometry resolution $[\frac{primitives}{m^2}]$ and 88 semantic classes. Habitat \cite{habitat19iccv} can render RGB, depth, semantic class segmentation images, and semantic instances at up to 10k frames per second.

\section{Matterport3D}
\begin{figure}[htpb!]
  \centering
    \includegraphics[width=0.9\textwidth]{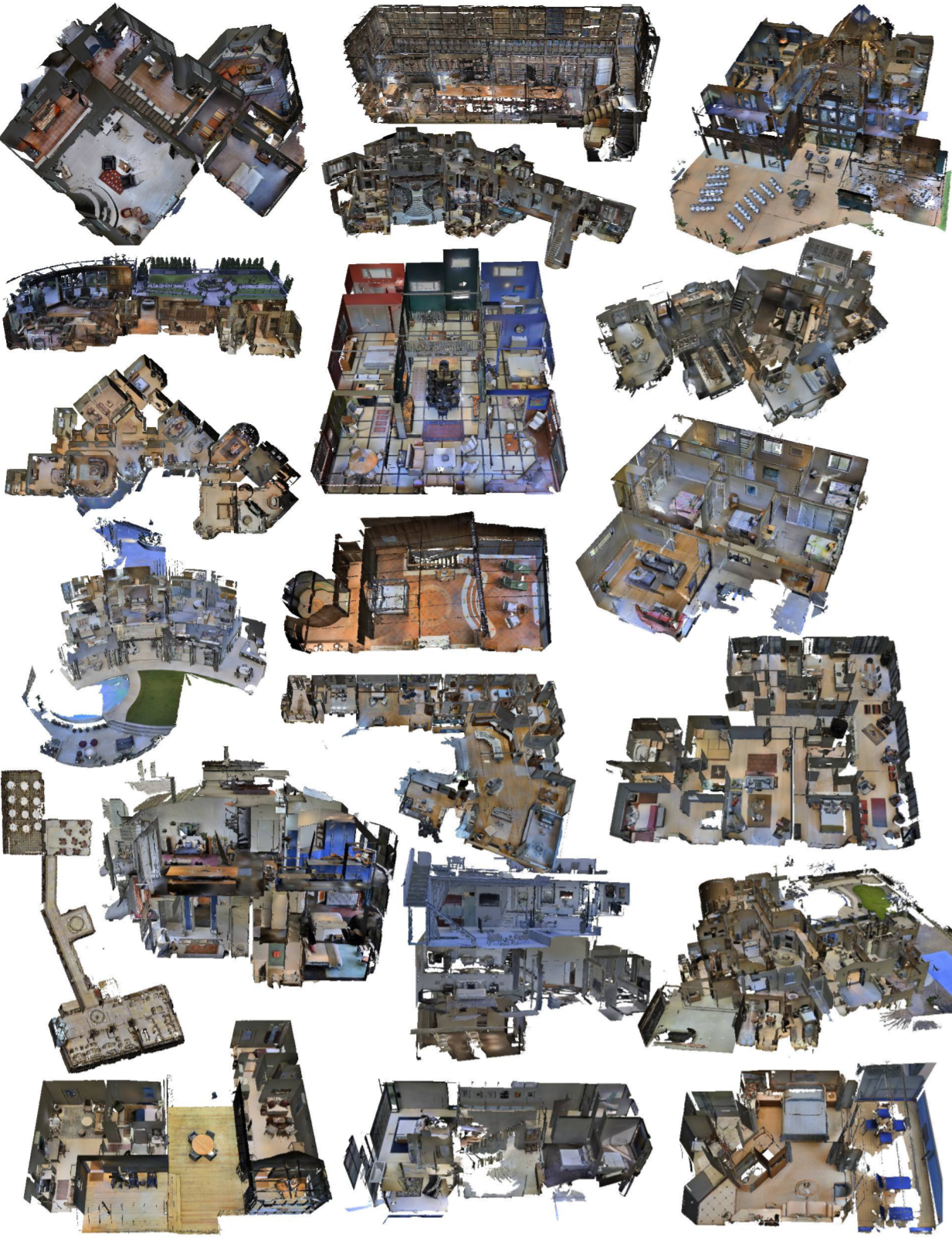}
    \caption{A set of different scenes from the Matterport3D dataset \cite{chang2017matterport3d}.}
    \label{fig:matterport3d}
\end{figure}

Matterport3D \cite{chang2017matterport3d} offers a highly diverse large scale RGB-D images for 90 building scenes, totalling in 10,800 panoramic views extracted from 194,400 RGB-D images. This dataset includes depth images, textured meshes, HDR color images, panoramic skyboxes,  objects semantic segmentation, region layouts, and categories. Each of these 90 scenes contains a group of RGB-D images at 1280x1024 (with color in HDR) captured with a 6 DoF camera pose estimate for each scene, a textured mesh for the entire scene, in addition to a skybox for each set of 18 images in the same panorama. It contains total of 2056 rooms with a higher color resolution of $97k$ $[\frac{pixel}{m^2}]$ compared to Replica \cite{straub2019replica}, but a lower geometry resolution of $0.7k$ $[\frac{primitives}{m^2}]$ and a lower number of semantic classes of 40 classes. Matterport3D is also a Habitat \cite{habitat19iccv} compatible; therefore, it offers a wide range of visual tasks for embodied agents to learn. Examples of Matterport3D scenes can be seen in Figure \ref{fig:matterport3d}.

\section{SoundSpaces} \label{soundspaces}
The aim of SoundSpaces is to enable the audio sensing capability to Habitat simulator \cite{habitat19iccv} by providing pre-computed binaural room impulse response (two channels to simulate how people receive the sounds with two ears)  for 85 scenes from Matterport3D \cite{chang2017matterport3d} and for the 18 scenes of Replica \cite{straub2019replica}. Since the Room Impulse Response (RIR) is defined as the transfer function between sound source and sound sensing device (Microphone), which varies depending on the sound source location, room geometry, and materials \cite{kuttruff2016room}, SoundSpaces considered these factors in order to provide a high fidelity audio simulator for realistic audio renderings in 3D environments. 
\par

SoundSpaces used the state-of-the-art algorithm for modeling of room acoustics from \cite{cao2016interactive}, modeled sound reflections in the room geometry by a bidirectional path tracing algorithm of \cite{veach1995bidirectional} and mapped the meshes’ semantic labels to materials in an existing database to include the acoustic material properties of major surfaces \cite{egan1989architectural} as each material has distinct scattering, absorption, and transmission coefficients that affect the sound propagation.
 \par

\begin{figure}
  \centering
    \includegraphics[width=0.9\textwidth]{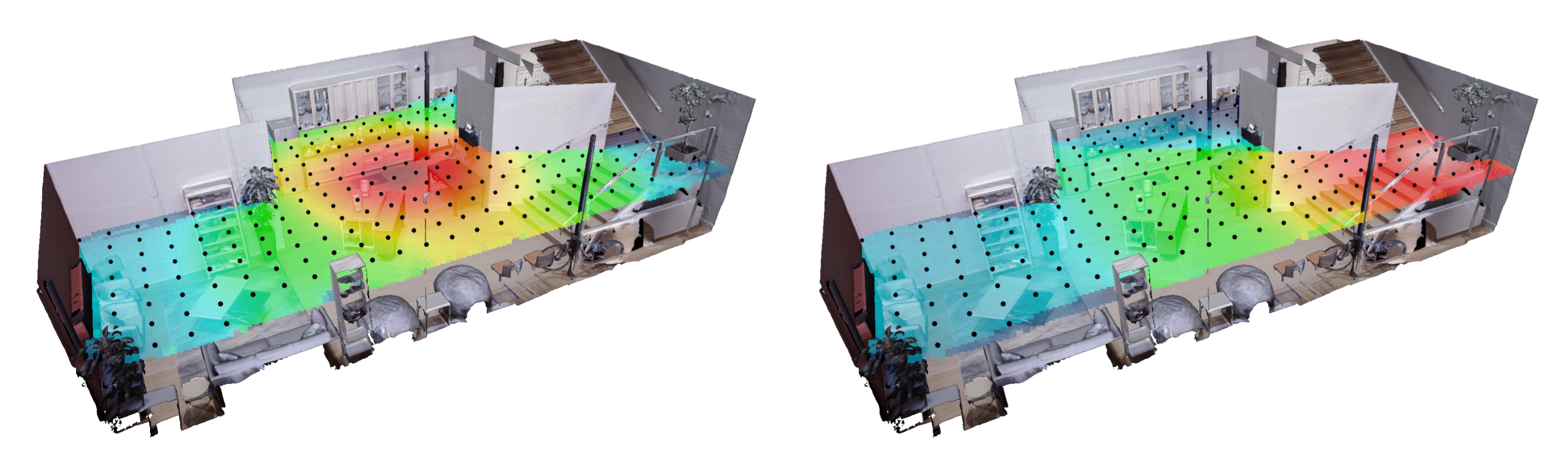}
    \caption{A grid map of one of the FRL apartments from the Replica dataset shows the locations of that scene's calculated binaural room impulse response. The heatmaps show audio pressure fields decreasing from red to blue. Left: when the sound source is located in the center. Right: when the sound source is located on the stairs \cite{chen2020soundspaces}.}
    \label{fig:soundspace}
\end{figure}

For calculating the Room Impulse Response (RIR) for each of the datasets, SoundSpaces densely sampled a grid of N locations with a spatial resolution of $1m$ for Matterport3D and $0.5m$ for Replica and since the area of Matterport3D scenes vary from 53.1 to 2921.3 $m^2$ thus N $\in [20,2103] $, while Replica scenes area varies from 9.5 to 141.5 $m^2$ with N $\in [38,566]$. All points are positioned at a vertical height of $1.5m$, representing the fixed height of an embodied robotic agent. After sampling the grid points, The Binaural Room Impulse Response (BRIR) has been calculated for each possible combination of listener and source locations by generating the ambisonic audio (a full-sphere representation of sound around the listener, it is like a 360-degree image of the sound) heard at every listener location for each different sound source location. This ambisonics converted to BRIR to provide the embodied agent with the sensing capability of the auditory events as If it has two human-like ears. SoundSpaces platform also provides a graph capturing the connectivity and the reachability of the sampled grid locations to prune the locations which are not navigable or not reachable by the agent due to the embodiment constraints. Figure \ref{fig:soundspace} shows how the audio pressure fields change across the grid locations. 
 

%% file: chapters/5-approach.tex
\chapter{Approach}
\label{chap:approach}
Current research on audio-visual navigation focuses on navigating towards a static AudioGoal without any distractors or noise \cite{chen2020soundspaces,gan2020look,chen2020learning}. While this provides a good start to tackle the AudioGoal navigation problem, current approaches suffer to generalize to unheard sounds or to show robustness to noisy observations or the presence of other sound sources. In this work, we increase the complexity of the current static audio-visual navigation task by covering a much broader range of scenarios. First, we introduce the novel dynamical audio-visual navigation task to tackle the scenarios with a moving sound-emitting source. Second, we integrate more complex real-world inspired scenarios by adding a second sound-emitting source, distractors, and noisy observations by performing feature-wise augmentation on the received audio signals. Finally, we propose a novel architecture to learn the two tasks of static and dynamic audio-visual navigation in the presence of complex audio scenarios.

\section{Problem Definition}
In this work, we tackle two problems with two setups for each. The first problem is the static AudioGoal task, in which the agent is required to navigate towards a static sound-emitting source in an unmapped 3D environment. While the second problem is the dynamical AudioGoal task, in which the target sound source is no longer static but moves in the environment. We use two setups for these problems, the first with clean audio environments that contain only the target sound without any noise, and in the second setup, we integrate more complex audio scenarios with more sound sources, noise, and disturbance, these complex scenarios are explained in more details in Section \ref{sec:complex}. The agent needs to rely only on audio and visual observations to navigate towards a previously unheard sound-emitting source in a previously unseen unmapped environment.
\par
We use Habitat simulator \cite{habitat19iccv} and its audio compatible SoundSpaces simulator \cite{chen2020soundspaces} which provides binaural room impulse response for Matterport3D \cite{chang2017matterport3d}, and Replica \cite{straub2019replica} datasets. Habitat and SoundSpaces provide a unique platform to train an embodied intelligent agent on the static and dynamic AudioGoal tasks using two publicly available highly photo-realistic datasets of Replica and Matterport3D, which consist of 3D scans of real indoor areas providing the agent with diverse experience of different real-world scenes with close and far AudioGoals. We use the same 102 copyright-free sounds from the static AudioGoal benchmark \cite{chen2020soundspaces} with the same train/val/test split 73/11/18 to train an intelligent agent on this benchmark and the novel dynamic audio-visual navigation benchmark. Detailed information about Replica, Matterport3D, and SoundSpaces is provided in the datasets Chapter \ref{chap:dataset}.
\par
Following the static AudioGoal task introduced by \cite{chen2020soundspaces}, The agent starts in a random position in an unknown environment in each episode. Then it has to navigate towards an unheard sound-emitting source location using actions of the Habitat simulator's discrete action space which consists of the following actions \textit{Move Forward, Rotate Left, Rotate Right}, and \textit{Stop}. The agent has to execute the \textit{Stop} action at the exact position of the sound-emitting source before the end of the 500 episode steps limit in order to gain the reward of the success. Following the waypoint selection approach \cite{chen2020learning}, the agent learns to select a waypoint instead of the low-level Habitat's actions as described in more details in the Action Parametrization Section \ref{sec:actions}.
\par
We tackle these problems using a reinforcement learning setup where the agent receives the current observation $o_t$ consisting of depth image $d_t$ and a binaural sound $b_t$ in the form of a spectrogram for the right and left ear.  The agent does not receive a displacement vector to indicate the goal position, making this task more challenging and different from the common PointGoal navigation task. The agent relies on this $o_t$ to navigate towards a previously unheard sound-emitting source in a previously unseen unmapped environment. Depending on the current observation $o_t$ and the state $s_{t}$, the agent then selects a next action $a_t$ from its policy $\pi(a_t | o_t, s_{t})$. The reinforcement learning agent aims to maximise the expected discounted return $\mathbb{E}_\pi[\sum_{t=1}^{T} \gamma^t r(s_{t}, a_t)]$, where $\gamma$ is the discount factor and $r(s_{t}, a_t)$ is the reward. 
\par
To compute the binaural spectrogram $b_t$ as in \cite{chen2020soundspaces, chen2020learning}, we convolve all sounds with the binaural room impulse response, which depends on the agent's current location and the sound source location, then for each channel, we take the magnitude of the output of the Short-Time Fourier Transform (STFT) with a hop length of 160 and a window length of 512 samples after padding with zeros and 400 before the padding. Then we downsample the output by a factor of 4; finally, we compute the logarithm of the output. After calculating the spectrogram for each channel, we combine the two channels together to form the binaural spectrogram $b_t$. Since the binaural room impulse responses provided by the SoundSpaces simulator \cite{chen2020soundspaces} have a different sampling rate of \SI{16000}{\hertz} for Matterport3D and \SI{44100}{\hertz} for Replica, the resulting binaural spectrogram $b_t$ shape differs between the two datasets. Matterport3D has a $b_t$ of size (65,26,2) and  Replica has a $b_t$ of size (65,69,2).
\section{Dynamic Audio-Visual Navigation}
\label{sec:dynamic_task}
\begin{figure}[htpb!]
  \centering
    \includegraphics[width=0.9\textwidth,keepaspectratio]{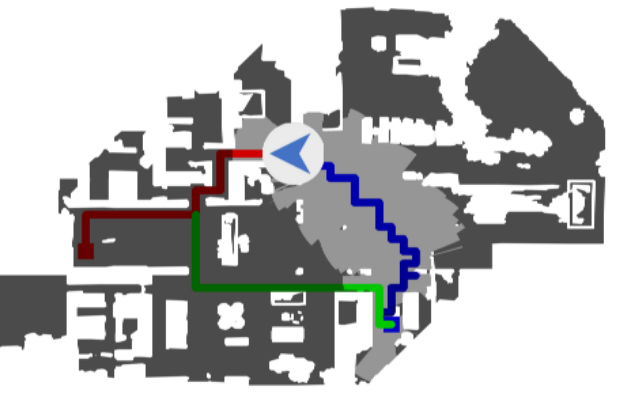}
    \caption{An illustration of the dynamic AudioGoal task, where the agent (blue square) aims to catch the moving target source (red square). The paths of the agent, sound source, and the optimal path are shown in blue, red, and green, respectively.}
    \label{fig:dspl}
\end{figure}
 In this task, the agent has to navigate towards a moving sound-emitting source in an unmapped complex 3D environment until it catches it and executes the \textit{Stop} action as explained earlier. The agent needs to reason not only about the current location of the moving sound but also the trajectory that the sound source takes to decide the earliest possible intersection point to reach the source at, depending on audio and visual observations. This setup increases the complexity of the existing static AudioGoal task to cover more challenging scenarios such as navigating to a person issuing commands or following a pet. In this setup, the agent has to update its memory more frequently as the previous observations no longer capture the current state of the environment as the target source keeps moving during the episode. This induces the agent to reason proactively about the trajectory path the moving source follows to catch it efficiently. To the best of our knowledge, this is the first work to tackle the problem of navigation to a moving sound-emitting source in unmapped complex 3D environments.

\subsection{Motion Model}
The target sound source starts in a random grid location in the environment and uniformly selects a goal location from the traversable grid, excluding the agent's current position, to avoid learning a sub-optimal policy where the agent waits until the moving source moves to its current location. The motion model also ensures the existence of a traversable path from its source's start position to the selected goal location. After choosing the target destination, the model generates a list containing the intermediate location the moving sound source has to pass by to follow the shortest path towards the selected goal destination. In each step, the model randomizes whether to move to the next grid node in that list or not with a probability of $30\%$. 
This percentage ensures the sound sources move slightly slower than the agent, ensuring that they can catch the moving sound source. Note that the moving source does not have an orientation and directly moves to the following location while the agent has to take separate rotation steps to change direction. Once the moving target reaches its goal, it draws a new random goal to navigate to; this motion model is preferable than to take a random movement every step, as this model adds reasonable and realistic motion behavior to the target source, which allows the agent to predict how it moves to learn to catch it at the earliest possible intersection point.

\subsection{Optimal Behavior: Dynamic Success weighted by Path Length (DSPL)}
\label{subsec:dspl}

The Success weighted by Path Length (SPL) \cite{anderson2018evaluation} serves as the primary metric to evaluate the navigation performance of embodied agents.  SPL measures the navigation performance for the successful episodes by calculating the ratio between the shortest path to the goal location and the path the agent took to reach that location. However, this metric relies on the prior knowledge of the shortest path to the goal location, which is not available in the newly introduced dynamic AudioGoal task as the goal location changes continually during the episode, and there is no prior knowledge of the trajectory the target source takes. Consequently, we introduce the Dynamic Success weighted by Path Length (DSPL) metric, which considers the shortest path length as the geodesic distance (the shortest traversable distance) between the agent's starting position and the earliest position the moving source passed by if the agent had enough steps from the beginning of the episode to catch the source in that position. Therefore, in every step, we calculate the number of steps the agent needs to catch the goal at the current goal position, and if this number is smaller than or equal to the current episode steps count, we consider the current target position as the earliest position the agent could catch the source at and the geodesic distance between the agent's start location and current goal location is considered as the shortest distance the agent needs to reach the goal. 

The DSPL is defined as follows:
\begin{equation}\label{3}
     DSPL = \frac{1}{N}\sum_{i=1}^{N}S_i \frac{g_i}{max(p_i,g_i)}.
\end{equation}
 where $i$ is the current episode count, $N$ is the total number of episodes, $S_i$ represents whether this episode is successful or not, $g_i$ is the shortest geodesic distance between the agent's start location and the closest position the agent could have caught the sound source at, and $p_i$ is the length of the path taken by the agent.\\
Note that this metric represents an oracle upper bound of the possible performance, which may not be achievable without a prior knowledge of the trajectory of the sound source. An example of the task and the optimal behavior as used in the DSPL is shown in Figure \ref{fig:dspl}.

\section{Complex Audio Scenarios}
\label{sec:complex}
\begin{figure}[htpb!]
  \centering
    \includegraphics[width=0.9\textwidth]{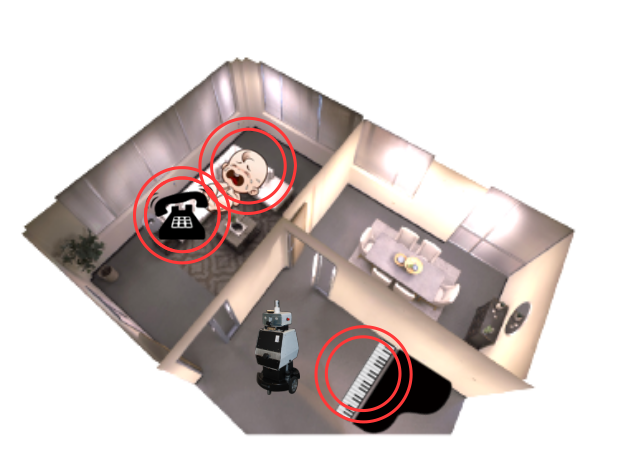}
    \caption{An illustration of one of the complex audio scenarios that we integrate. The agent is required to navigate towards the ringing phone in the presence of a second sound source at the same location (a crying baby), and other distractor sounds like a piano \cite{younes2021catch}.}
    \label{fig:complex}
\end{figure}
\input{tables/randomization_algo}
The ultimate goal of training embodied AI agents to navigate to a sound-emitting source is to deploy them in real-life to perform some of the tasks that require navigating towards a sound source (e.g., doorbell, telephone). However, 
current approaches  \cite{chen2020soundspaces,gan2020look, chen2020learning} primarily focus on relatively simple scenarios with a single sound-emitting source in a clean audio environment. In contrast, the explorations in noisy environments with sound disturbances are still minimal, as discussed in the related work Chapter \ref{chap:relatedwork}.
To mimic some challenging real-world scenarios and provide the agent with a more realistic, highly diverse training experience, we confront the agent with designed complex audio scenarios by including second sound-emitting sounds at the goal location, distractors sounds at different locations, and noisy audio sensors. These scenarios encourage the agent to focus on the spatial and directional understanding ingrained in the audio signal to improve the generalization to unheard sounds in noisy environments at test time.
\par
We introduce a randomization pipeline to provide the agent with highly randomized complex audio scenarios with perturbations and augmentations on episode and step levels. The pipeline randomizes the insertion of the following three main components:
\begin{itemize}
    \item Second Sound Source: For each episode, with a certain probability, the pipeline includes an additional different audio signal coming from a second sound source located at the same position as the target sound.
    
    \item Distractor Sounds: The pipeline includes distractor sounds in each episode with a fixed probability. If the episode includes distractors, the pipeline further randomizes whether the distractor is audible in each step, and if it is audible, a random distractor sound is chosen from the available training sounds, excluding the current target sound. Then the pipeline selects a random location to insert this distractor sound in the environment at, excluding the current target position.
    
    \item Spectrogram Augmentations: In each step, a random augmentation is applied to the spectrogram. We adopted two of the SpecAugment components introduced by \cite{park2019specaugment} to construct a set of augmentations (none, time masking, frequency masking, both). These augmentations are explained in more details in Section \ref{sec:specaugment}.
\end{itemize}

On the one hand, these scenarios increase the complexity of the task, demanding the agent to reason about the components of the audio signal. At the same time, it tremendously increases the diversity of training experience, which has shown to be very beneficial, especially in scenarios with limited data, such as the relatively small audio dataset of 102 sounds used here. All augmentations are solely based on the training sound dataset to ensure that the agent does not hear any sounds from the validation or test sets. The entire randomization pipeline is explained in Algorithm \ref{alg:algorithm1} and an illustration of one of these complex scenarios is provided in Figure \ref{fig:complex}.

\section{Dynamic Audio Visual Navigation (DAV-Nav) Model}

\begin{figure}[htpb!]
  \centering
    \includegraphics[width=0.9\textwidth]{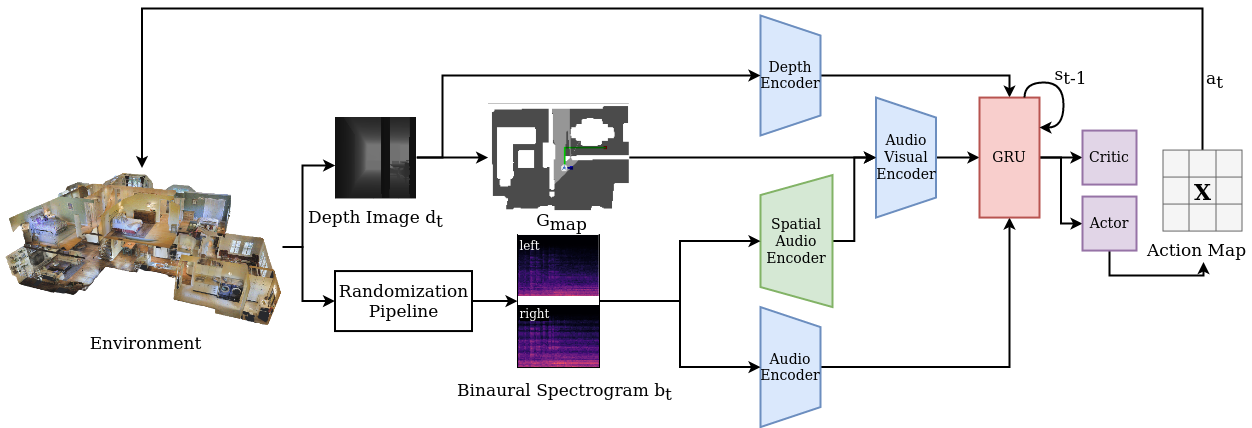}
    \caption{Our architecture. The depth image is projected into an occupancy geometric map $G_{map}$. Audio-Visual Encoder combine the learned spatial features from the Audio Spatial Encoder and the geometric map. The GRU combines these encoded features with the previously learned features. The PPO agent then outputs close-by waypoints that are executed by a Djikstra planner.}
    \label{fig:arch}
\end{figure}

Current research on audio-visual navigation focuses on straightforward end-to-end reinforcement learning from visual and audio inputs to actions. AV-Nav \cite{chen2020soundspaces} encodes binaural spectrogram of audio observation and RGB-D images independently, while AV-WaN \cite{chen2020learning} additionally uses the audio and visual inputs to construct acoustic and geometric maps before encoding them separately. Then both approaches combine the individual encoded features and depend on a standard Gated Recurrent Unit (GRU) cell  \cite{cho2014learning} to combine them with previously learned features. While the work of  \cite{gan2020look}  uses the audio signal to estimate the relative goal location and estimates a local occupancy map, then combines both inputs and passes them to a planner to execute the needed actions to reach this estimated goal. Nevertheless, none of these previous approaches provide a neural network architectural component to learn a combination of these audio and visual modalities.

Binaurally received spectrograms of the audio source retain a large amount of information about the room geometry and space due to the result of reflections of sound propagation through the rooms, objects, and walls. Prior work has shown that this information can reveal room geometries and can accelerate the floor plan reconstruction by a large margin compared to methods that rely on visual perception only \cite{purushwalkam2020audio}.
In this work, we hypothesize that learning to extract and focus on this embedded information in the received sound observations and learning a proper way to combine this information with the spatial information from occupancy maps is an advantageous architectural prior for AudioGoal navigation tasks. Moreover, we hypothesize that a neural network architectural component that captures this aspect of the sound information has a higher likelihood of generalizing to unheard sounds and more robust to distractors and noisy audio environments.

Our model receives binaural spectrogram $b_t$ and depth image $d_t$ as input observations. Then similarly to AV-WaN, the model constructs and continuously updates an allocentric geometric map $G_{map}$ from the depth inputs $d_t$. This map $G_{map}$ consists of two channels, one for occupied/free space and the other one for explored/unexplored areas.
Then the model passes the  $b_t$ observation to two encoders, namely an Audio Encoder Section \ref{subsec:audio} and a Spatial Audio Encoder Section \ref{subsec:spatial}. The first one is responsible for learning direct useful information from the observation, while the latter one maps the binaural spectrogram into a spatial feature space. The architecture also has a Depth Encoder Section \ref{subsec:depth} that directly encodes the visual observation. The Audio Visual Encoder Section \ref{subsec:audio-visual} proposes an early fusion of the learned spatial audio features and the spatial geometric map $G_{map}$ based on convolutional layers. Similar to AV-Nav and AV-WaN, the model then uses a Gated Recurrent Unit (GRU) as a memory component and trains the agent end-to-end with Proximal Policy Optimization (PPO) \cite{schulman2017proximal}. The PPO algorithm is explained in more details in Section \ref{sec:ppo} while the GRU is explained in Section \ref{sec:gru}. We provide the implementation details of all models in the following subsections and an illustration of our approach in Figure \ref{fig:arch}.

\subsection{Action Parametrization}
\label{sec:actions}
The AV-Nav \cite{chen2020soundspaces} baseline reasons directly in Habitat's raw action space, while the AV-WaN \cite{chen2020learning} baseline demonstrated the benefits from learning on a higher-level action space as its model learns to select intermediate waypoints. The AV-WaN baseline uses a $9\times9$ action map centered on the agent's current location to select a waypoint from it, then passes this waypoint to the Dijkstra planner, which executes the low-level required actions to navigate to the chosen waypoint. Although acting in such higher-level action space can be helpful, selected far-away waypoints decrease the control frequency as it takes a longer time to reach, which results in ignoring up to ten observations. The loss of information from these observations becomes more critical in our highly dynamic setups as the observations that the agent used to select the waypoint might not present the current state of the environment anymore due to the moving of the sound source or the existence of distractors and noisy observations.
Therefore, we decrease the size decrease of the action map to $3\times3$ to get an adequate middle ground between the advantages of learning waypoints and limiting the number of ignored observations to a maximum of four.

\subsection{Reward}

The agent receives a positive reward of +10 when it successfully executes the \textit{stop} action at the exact location of the sound source. It further receives a small dense reward of +0.25 for decreasing and -0.25 for increasing the shortest path distance to the goal. 
For the moving sound source task setup, the shortest distance is between the agent and the current sound source's location. This shortest distance differs from the shortest distance to the earliest possible reachable intersection location used to calculate the DSPL as explained in this Section \ref{subsec:dspl}, which is the primary metric for evaluating this novel task; this decreases the contribution of the dense supervisory reward signal to direct the agent to learn the optimal policy.
Finally, a small-time penalty of -0.01 for each step discourages the agent from taking unnecessarily rotation actions at the same location.

\subsection{Depth Encoder}
\label{subsec:depth}
We adapt the depth encoder introduced by AV-Nav~\cite{chen2020soundspaces} baseline. The encoder has three convolutional (CNN) layers with kernel sizes of (8, 8), (4, 4), and (3, 3), and with strides (4, 4), (2, 2), and (2,2). There is a ReLU activation function after each CNN layer and a fully connected layer at the end of the encoder then embeds the features to a size of 512, followed by a ReLU activation function. 

\subsection{Audio Encoder}
\label{subsec:audio}
Based on the architecture of \cite{chen2020learning}, the model encodes the audio signals with three convolutional layers with kernel sizes of (8, 8), (4, 4) and (3, 3) and strides of (4, 4), (2, 2) and (1, 1) for Replica~\cite{straub2019replica}, and kernel sizes of (5, 5), (3, 3), and (3, 3) and strides of (2, 2), (2, 2), and (1, 1) for Matterport3D~\cite{chang2017matterport3d} datasets. There is a ReLU activation function after each convolution layer and a fully connected layer at the end of the encoder then embeds the features to a size of 512, followed by a ReLU activation function

\subsection{Spatial Audio Encoder}
\label{subsec:spatial}
This encoder receives as an input the binaural spectrogram $b_t$ of size (2, 65, 69) for Replica and (2, 65, 26) for the Matterport3D datasets. The Spatial Audio Encoder upscales the input to the same dimensionality as the geometrical map (2, 200, 200). Due to the different spectrogram dimensionality of the two datasets, the model has two different encoders. Replica's encoder consists of two transposed convolution layers with kernel sizes of (8,8) and (1,13), and with strides of (3,3) and (1,1), respectively. The encoder uses a ReLU activation function after each convolution layer. For Matterport3D, the encoder consists of two transposed convolution layers followed by one convolution layer. The kernel sizes are (5,2), (4,2) and (1,5) while the strides sizes are (3,4), (1,2) and (1,1), respectively. The encoder also uses a ReLU activation function after each convolution layer.

\subsection{Audio-Visual Encoder}
\label{subsec:audio-visual}
The model in Figure \ref{fig:arch} concatenates the output of the Spatial Audio Encoder with the current geometrical map $G_{map}$. Subsequently, it passes the resulting features through three convolutional layers with ReLU activations, followed by a fully connected layer to embed the features into an embedding of size 512. Finally, the encoder uses a ReLU activation function and feeds the output to a GRU. The kernel sizes are (8, 8), (4, 4), and (3, 3), while the strides are (4, 4), (2, 2), and (1, 1).

\subsection{Actor-Critic}

We implement the Gated Recurrent Unit (GRU) \cite{cho2014learning}  as a single bidirectional GRU cell \cite{chung2015recurrent} with hidden size 512 and one recurrent layer.
The actor and critic take the output current state of the GRU and estimate the action distribution $\pi(a_t|s_t)$ and value of the state $v(s_t)$, respectively. We implement both as a linear layer. We use a categorical distribution with probabilities based on the softmax of the actor's output to select the waypoint. We also add an entropy maximization term for exploration \cite{haarnoja2018soft} to the Proximal Policy Optimization (PPO) \cite{schulman2017proximal} objective like \cite{chen2020soundspaces, chen2020learning}.

%% file: tables/randomization_algo.tex
\begin{algorithm}

\textbf{Require: }{listOfSounds: training sounds excluding current episode target sound, targetAudio: current episode target sound, agentPosition, targetPosition, computeAudio: function to compute audio observation, computeSpectrogram: function to compute spectrogram, applySpecAugment: function to apply feature augmentation, listOfNodes: traversable grid locations, rnd: uniformly random choice function.}
\newline
\SetAlgoLined
\For{episode in episodes}
{
includeSecondSound $=$ rnd([True, False])

\If{includeSecondSound}{
secondAudio $=$ rnd(listOfSounds)}

includeDistractor $=$ rnd([True, False])

\For{step in steps}{
audio $=$ computeAudio(targetAudio, agentPosition,targetPosition)
\\
\If{includeSecondSound}{
audio  +$=$ computeAudio(secondAudio, agentPosition, targetPosition)
}
distractorStep $=$ rnd([True, False])

\If{includeDistractor \textbf{and} {distractorStep} }{
distractorAudio $=$ rnd(listOfSounds)
\\
distractorPosition $=$ rnd(listOfNodes)
\\
audio +$=$ computeAudio(distractorAudio $,$ agentPosition$,$ distractorPosition)
}
spectrogram $=$ computeSpectrogram(audio)
\\
augment $=$ rnd([True, False])
\\
\If{augment}{
aug  $=$ rnd([timeMasking, frequencyMasking, both])
\\
applySpecAugment(spectrogram, aug)
}
}
}
 \caption{Randomization Pipeline}
 \label{alg:algorithm1}
\end{algorithm}

%% file: chapters/6-experiments.tex
\chapter{Experimental Evaluation}
\label{chap:experiments}

\section{Task Setups}

In this thesis, we tackle the tasks of static and dynamic AudioGoal navigation. For each task, we train all agents on two setups: the same clean audio setup used in \cite{chen2020soundspaces, chen2020learning} and the proposed complex audio scenarios with noisy audio, distractors, and second sound-emitting source as introduced in Section \ref{sec:complex}. All agents are trained on multiple sounds and evaluated in two settings: on heard sounds in unseen environments and unheard sounds in unseen environments. We use the same train/val/test splits protocol used by \cite{chen2020soundspaces, chen2021learning}, where Replica splits into 9/4/5 scenes, and Matterport3D (MP3D) splits into 59/10/12 scenes. The 102 different sounds split into 73/11/18. We apply the same split to other audio signals, such as distractors. This means that the agents will confront unheard distractor sounds during the evaluation while navigating to previously unheard sound sources.  

\section{Metrics}
We use the following metrics to evaluate the navigation performance:
\begin{itemize}
\item \textit{Success rate (SR)}: The percentage of successful episodes of all test episodes. We consider an episode successful if the agent executes the $stop$ action at the exact target goal location.
\item \textit{Success weighted by path length (SPL)} \cite{anderson2018evaluation}: This metric calculates the ratio of the length of the shortest path to the goal to the length of the executed path for the successful episodes.
\item \textit{Success weighted by number of actions (SNA)} \cite{chen2020learning}: The ratio of the number of actions needed to follow the shortest path to the actual number of actions the agents took to reach the same goal. In contrast to the SPL, this metric considers the number of orientation changes.
\item \textit{Dynamic success weighted by path length (DSPL)}: The primary metric for the novel task of dynamic AudioGoal Navigation. It presents the ratio of the length of the path to the earliest possible reachable intersection location and the length of the actual executed path for the successful episodes. An illustration of the DSPL metric is presented in Figure \ref{fig:dspl}.
\item \textit{Dynamic success weighted by number of actions (DSNA)}:
Equivalently to the DSPL, the adjusted version of the SNA calculates the ratio of the number of actions needed to follow the shortest path to reach the earliest possible reachable intersection location to the actual number of actions the agents took to reach the goal.
\end{itemize}

\section{Baselines} 
To evaluate our proposed approach, we compare it with the two current state-of-the-art methods: AV-Nav \cite{chen2020soundspaces}, and AV-WaN \cite{chen2020learning}. \textit{AV-Nav} \cite{chen2020soundspaces} is an end-to-end reinforcement learning agent that directly encodes visual and  audio observations to select actions from Habitat's \cite{habitat19iccv} discrete action space.  We use the authors' code to train and evaluate the model. \textit{AV-WaN } \cite{chen2020learning} is the current state-of-the-art for static AudioGoal task, which predicts intermediate waypoints to the goal depending on audio observation, acoustic, and geometric maps. We use the code provided by the authors for training and evaluating this baseline. 

\section{Static AudioGoal Task}
\input{tables/static_without_complex}

Starting with evaluating the agent's performance on the static AudioGoal task, we trained the agents on multiple sounds and evaluated them in unseen environments with either the heard or unheard sounds. We show the results for the original setup without including the proposed complex audio scenarios in the upper half of Table \ref{tab:static}. The proposed novel architecture that learns the spatial fusion between the two modalities (visual and audio) achieves a significantly improved performance on unheard sounds setup while performing similar or even better on the heard sounds setup compared to the baselines. Our model improved the success rate from 51.1 to 63.6 percent on Replica and 56.6 to 60.6 percent on Matterport3D, with comparable improvements on SPL and SNA metrics.

We show the effects of training on the complex audio scenarios described in Section  \ref{sec:complex} and evaluation on the standard benchmark in the bottom part of the table. The results demonstrate that training on these scenarios provides all models with better generalization performance to unheard sounds, with above 30 percentage points for specific models. Furthermore, our proposed model achieves the best performance across all metrics. Combining the new architecture with the complex audio scenarios improves the navigation performance from an SPL of 27.3 to 54.0 on Replica and 39.6 to 55.0 on Matterport3D, strongly outperforming prior state-of-the-art results on this setup.
We provide a decomposition of the effects of the individual components of the complex scenarios in Section \ref{sec:decomp}.

Training with the complex scenarios and evaluating on the heard sounds setup only in Replica shows a drop in performance of our proposed architecture. This can be due to the small areas of Replica as they vary from 9.5 to 141.5 $m^2$ or due to feature-wise augmentation that may mask specific channels that the model could previously exploit to overfit on heard sounds setup. \\
Then we conducted a further evaluation on the impact of including dynamic target sounds within the training. In each episode, the randomization pipeline randomly selects whether to include a dynamic or static sound to test the model's ability to learn both tasks simultaneously. Although training this setup on Replica came with a negative impact, performance on Matterport3D did not change. Replica's smaller scenes areas could be the reason for this difference. Proceeding to the evaluation of performance on the complex audio scenarios, the results as shown in Table \ref{tab:static_complex} show lower performance for our models and the baselines, which demonstrate that these scenarios increase the complexity of the task. We observe that the decrease in performance is much smaller on the large Matterport3D dataset, with similar performance on unheard sounds compared to the non-complex scenarios setup. The results of our model in the presence of complex audio scenarios demonstrate the generalization capability of our proposed model on unheard sounds.
\input{tables/static_with_complex}

\section{Dynamic AudioGoal Task}
\begin{figure}[htpb!]
  \centering
    \includegraphics[width=0.9\textwidth,height=0.9\textheight, keepaspectratio]{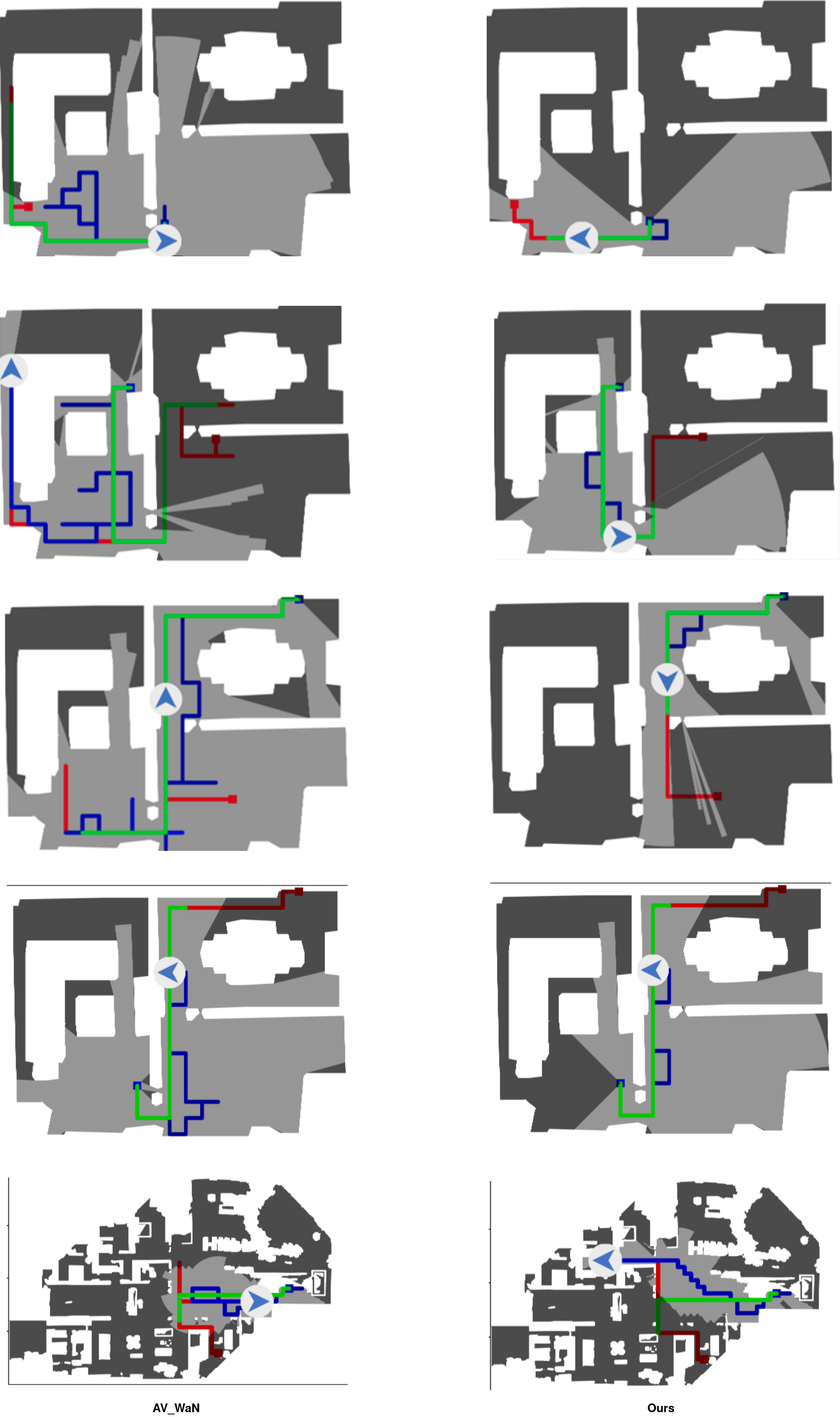}
    \caption{Navigation performance comparison between the Av-WaN baseline \cite{chen2020learning} (left) and our proposed architecture (right) on the dynamical audio-visual navigation task. The agent (blue square) aims to catch the moving target source (red square). The paths of the agent, sound source, and the optimal path are shown in blue, red, and green, respectively.}
    \label{fig:examples}
\end{figure}

\input{tables/dynamic_without_complex}
Moving to the evaluation of the dynamic AudioGoal task, we trained all models on multiple sounds, with a moving sound source for 50\% of the episodes and with a stationary sound source for the other 50\%. Then we evaluated all models on the dynamic AudioGoal task for both heard and unheard sounds setups. Results as shown in Table \ref{tab:dynamic} show the ability of all models to solve the task, with success rates (SR) of 69.5-95 percent on heard sounds which is a comparable performance to the static AudioGoal task. However, there is a broader gap in performance on unheard sounds setup in terms of the success rates (SR) and the optimality of agent's paths.
Our proposed architecture distinctly outperforms the baselines on all clean audio setups except for slight advantages for AV-WaN on the Replica dataset on unheard sounds setup.
The results again affirm that training on the complex audio scenarios is advantageous for the generalization to unheard sounds for all models except AV-Nav. This can be due to the reasoning directly in the raw low-level action space, making the task harder for the agent as it does not clearly understand which sound source it shall follow. Again, our proposed architecture trained with these complex scenarios achieves substantial advancements from a success rate of 38.7 for the best baseline on clean audio scenarios to 68.0 on Replica and from 38.7 to 74.0 on Matterport3D. The results of the same models evaluated on the complex audio scenarios are shown in Table \ref{tab:dynamic_complex}. While on Replica, the best result alters depending on the metric between AV-WaN and our proposed architecture, while on Matterport3D, our proposed model consistently performs the best. We included some episodes examples in Figure \ref{fig:examples} that show the navigation performance of our model compared to the navigation performance for the best baseline.
\input{tables/dynamic_with_complex}

\section{Hyperparameter Optimization}

In this section, we explain the important Hyperparameter optimization experiments we conducted.
\subsection{Spectrogram Augmentation Components}
The work of SpecAugment as introduced by \cite{park2019specaugment} and explained in Section \ref{sec:specaugment} has three types of spectrogram augmentations on every data point. It uses 1. Time Wrapping. 2. Time Masking. 3. Frequency Masking. However, we conducted several experiments on every component independently, and we found that the best method to provide the agent with a highly diverse experience is to randomize the selection of augmentation type in every environment step from this set [Time Masking, Frequency Masking, Both, None].
We dropped Time Wrapping from the selection as it was the least efficient and the most expensive in terms of execution time as reported in \cite{park2019specaugment}.

\subsection{Time and Frequency Masking}

\begin{figure}[htpb!]
  \centering
    \includegraphics[width=0.9\textwidth,keepaspectratio]{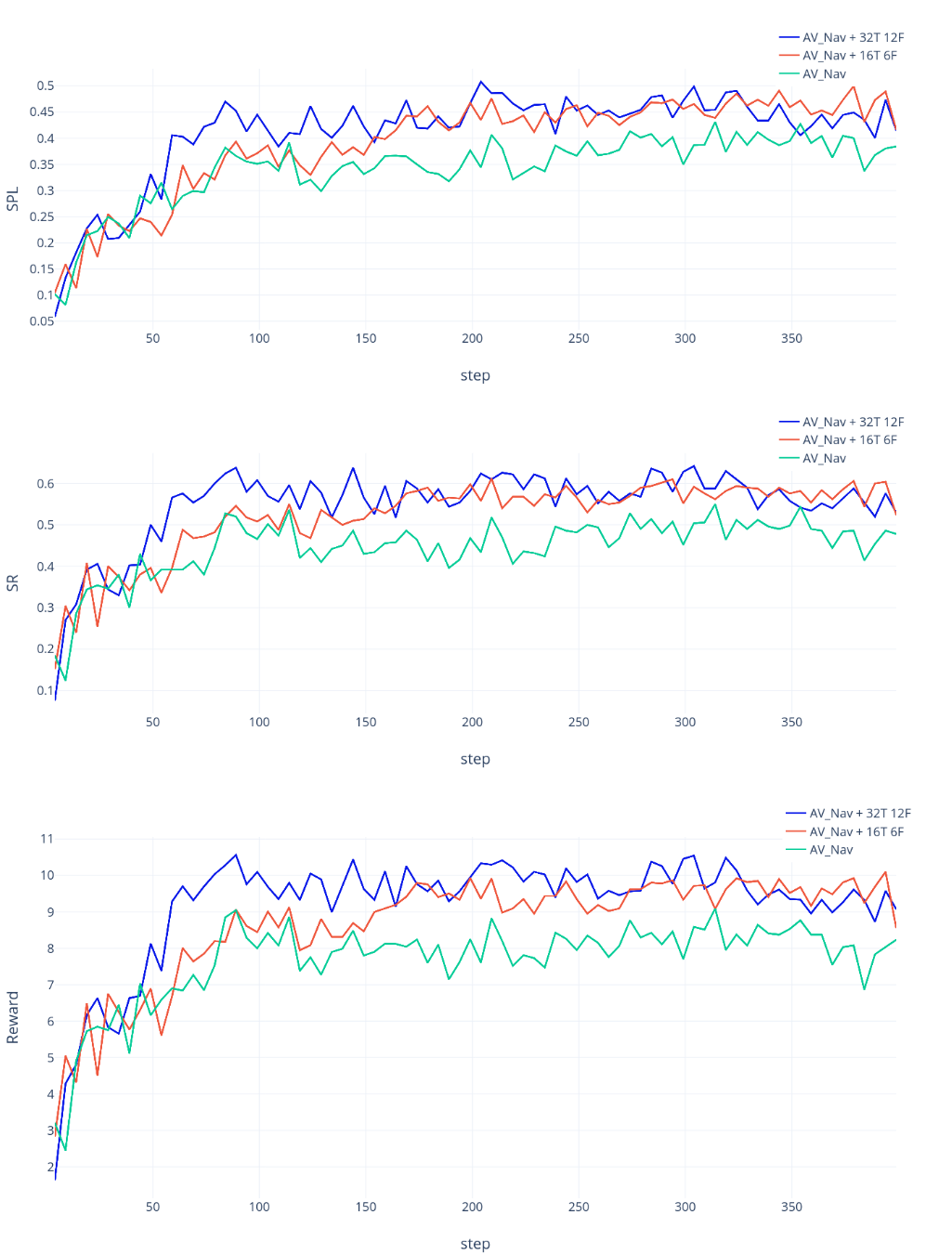}
    \caption{A comparison between the AV-Nav baseline \cite{chen2020soundspaces} and the best two combinations of Time and Frequency Masking parameters on Replica. The metrics from top to bottom are SPL, SR, and Reward. The baseline is in green, while the best combination is in blue, followed by the second-best combination in red.}
    \label{fig:replica_spec}
\end{figure}
\begin{figure}[htpb!]
  \centering
    \includegraphics[width=0.9\textwidth,keepaspectratio]{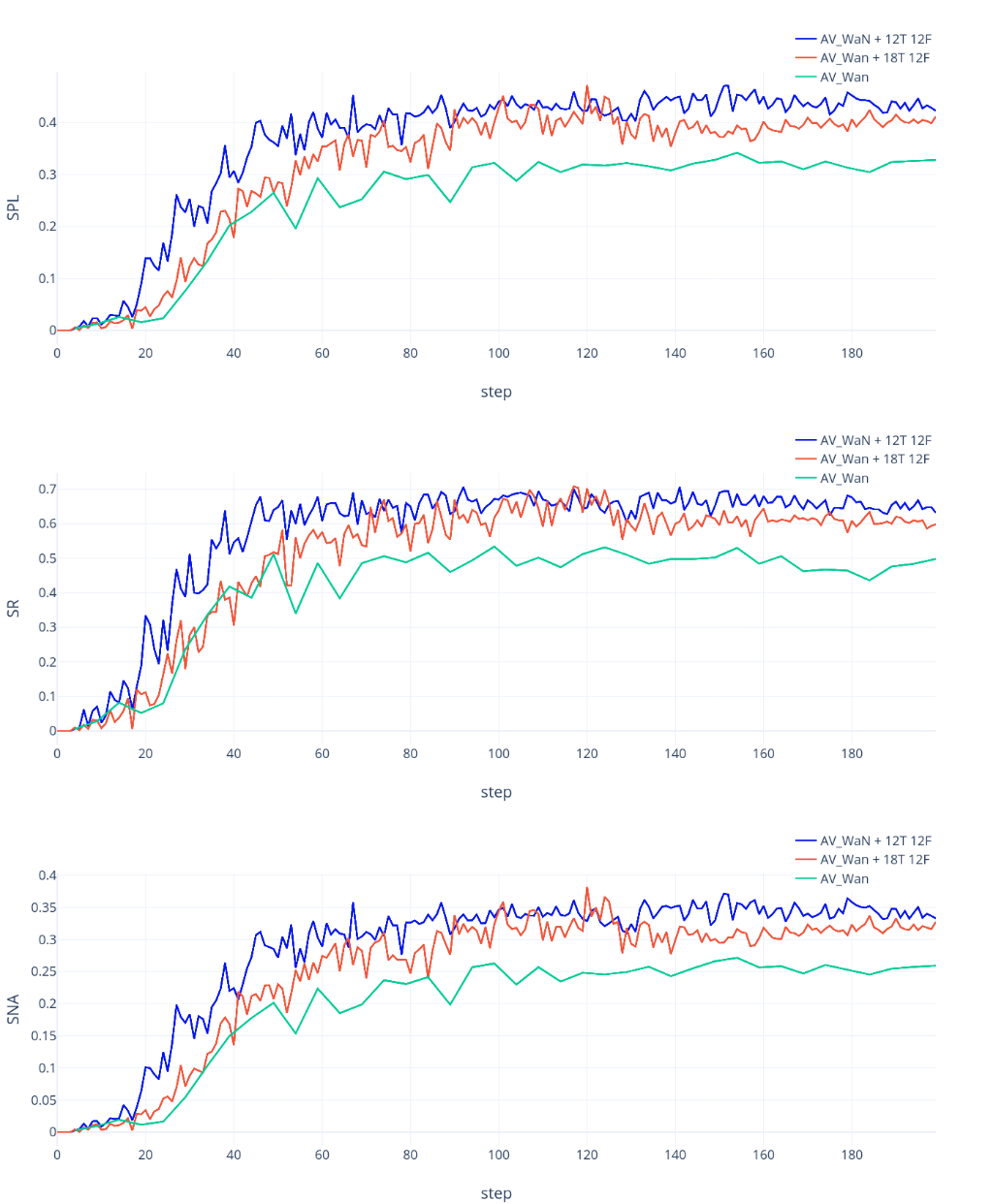}
    \caption{A comparison between the AV-WaN baseline \cite{chen2020learning} and the best two combinations of Time and Frequency Masking parameters on MP3D. The metrics from top to bottom are SPL, SR, and SNA. The baseline is in green, while the best combination is in blue, followed by the second-best combination in red.}
    \label{fig:mp3d_spec}
\end{figure}

The selection of the value of parameters of time and frequency masks' is very crucial as it has to find a middle ground between 1. Injecting more noise to induce the agent to learn the important features without overfitting the current dataset. 2.  The observation does not lose its vital features, which will ruin the training. Due to that, we conducted several experiments with different frequency and time masks' values on both datasets as each of them has a different spectrogram shape.
On Replica, we found that the best combination is (Frequency Masking Parameter = 12, Time Masking Parameter = 32) as shown in Figure \ref{fig:replica_spec}, while on Matterport3D, the best combination is (Frequency Masking Parameter = 12, Time Masking Parameter = 12) as shown in Figure \ref{fig:mp3d_spec}.
\subsection{Complexity of Audio Scenarios}
\label{sec:decomp}
In this section, we investigate the importance of the individual elements of the audio scenarios. The results in Table \ref{tab:decomp} show the performance for models trained with subsets of these elements on the static AudioGoal task. On unheard sounds setup, training with the second audio source at the exact goal location improves the performance on both datasets, while the randomized spectrogram augmentations improve generalization on the Replica dataset, but not on Matterport3D. While combining second audio sources and spectrogram augmentations improves the overall performance on the Matterport3D dataset but not on Replica. Finally, the integration of distractor sound sources further improves the generalization on Matterport3D but does not have a large impact on Replica. The variations in the individual complex audio scenarios elements' impact across Replica and Matterport3D datasets show the importance of combining different perturbations and randomizations to provide the agent with a highly diverse experience to learn valuable features instead of overfitting to a custom set of sounds.

On the heard sounds setup, disturbances and perturbations from the complex audio scenarios have a negligible effect on learning to navigate to heard sounds in unseen environments on the Matterport3D dataset. While on Replica, the decomposition of complex components shows that the performance's drop was not connected to a single element but with any disturbance. This might signify that training the agent without the proposed complex scenarios increases the possibility of learning features of the audio signal that do not help in generalization or robustness to perform this task in much smaller environments. In addition to that, the results show the impact of using these complex scenarios might be larger as the distances to the goals are generally shorter, as that provides the agent with a fewer number of audio observations to filter out such disturbances.

The results of the evaluation with complex scenarios at test time as shown in Table \ref{tab:decomp_complex} demonstrate that training with all complex audio scenarios components improves robustness in performing the task in the presence of these real-world inspired scenarios. There is no subset of the components found that provides the same robustness to the full set of perturbations. Moreover, the performance of the model trained on clean audio without the proposed audio complex scenarios drops in most of the cases more than 50\% across all metrics and settings, which shows the importance of including the proposed scenarios in training to achieve a robust behavior in complex and noisy audio environments.

\input{tables/augmentation_decomposition}

\subsection{Action Map Size}
\begin{figure}[htpb!]
  \centering
    \includegraphics[width=0.9\textwidth,keepaspectratio]{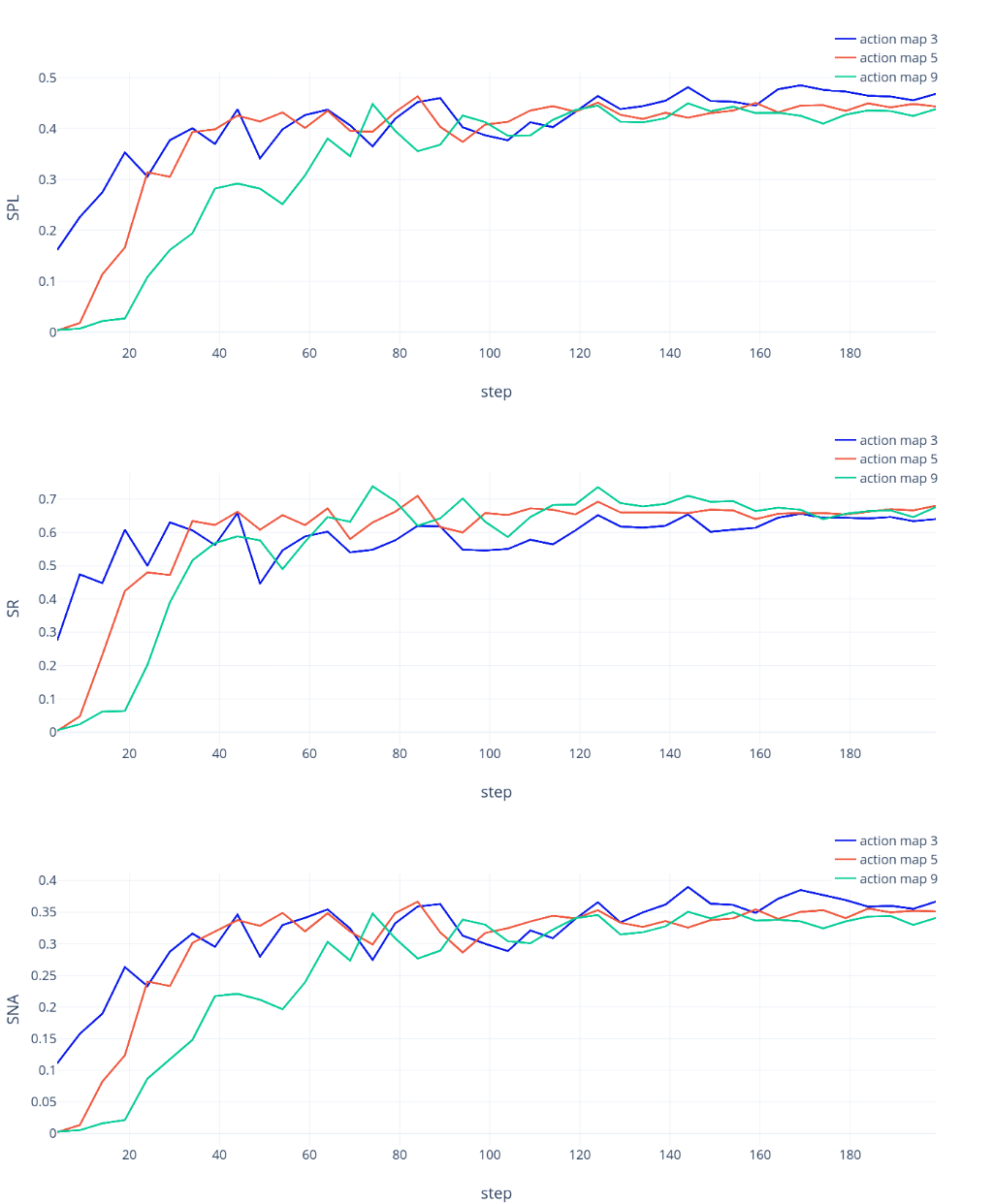}
    \caption{A comparison between different action maps size ($9\times9$, $5\times5$ and $3\times3$). The metrics from top to bottom are SPL, SR, and SNA. In terms of overall performance, we found that the $3\times3$ action map size helps the agent to learn the task faster.}
    \label{fig:action_map}
\end{figure}
{The AV-WaN \cite{chen2020learning} uses an action map of size $9\times9$; however, after trying an action map of size $5\times5$ and $3\times3$, we found that decreasing the action map size to $3\times3$ speed up the learning. Decreasing the action map size makes it easier for the agent to learn a smaller set of surrounding waypoints to choose from in every step; in addition to that, it helps it to recover quicker from selecting a bad waypoint. This decrease in action map size became much more beneficial for the setup with the complex audio scenarios and the setup of novel dynamic audio-visual navigation task with or without complex scenarios. These setups require the agent to be more conscious of any change in the environment; therefore, the agent should not rely much on the previous observations as it may not be a good representation of the current state of the environment anymore.}
\include{tables/hpo}
In Table \ref{tab:hpo}, we represent a summary of most of the hyperparameter optimization experiments that we conducted, while in Table \ref{tab:hyper}, we show the final hyperparameters used for our model and the baselines.
\input{tables/hyperparameters}

\section{Further Ablations}
\subsection{Spectrogram Reconstruction}
\label{sec:aux}
We observed that the models struggle to generalize to unheard sounds, so we decided to integrate an extra supervisory signal to ensure that the model learns the vital features instead of overfitting to the heard sounds. We proposed a new auxiliary task for this purpose that aims to reconstruct the spectrogram from the encoded features, so we make sure that these features cover all crucial information that is essential to provide a good representation for the audio observation. This auxiliary task is responsible for inducing the Audio Encoder Section \ref{subsec:audio} to extract a better-encoded feature space. The auxiliary task tries to reconstruct the binaural spectrogram $b_t$ using an Audio Decoder by passing to it the extracted features from the Audio Encoder. Then we calculate an additional Mean Squared Error Loss between the current binaural spectrogram $b_t$ and the reconstructed one $b_t^{'}$ using Equation \ref{eq:aux_loss}  and pass the calculated loss to the total loss after weighting it to match the same order of magnitude. 
\begin{figure}[htpb!]
  \centering
    \includegraphics[width=\textwidth,keepaspectratio]{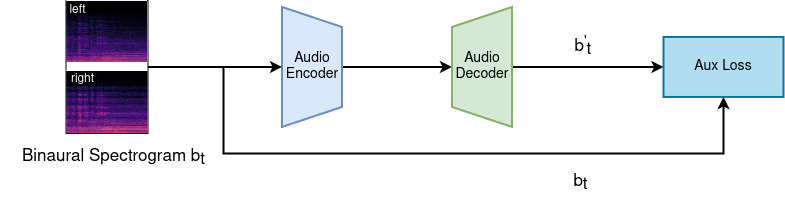}
    \caption{Reconstruction of spectrogram auxiliary task architecture.}
    \label{fig:aux_task}
\end{figure}
\begin{equation}
     auxiliary\: loss = \frac{1}{N} \sum_{i=1}^{N} (b_{t_i} - b_{t_i}^{'})^2
    \label{eq:aux_loss}
\end{equation}
where $N$ is the number of the running parallel environments.
\subsection{Auxiliary Task Evaluation}
Although the spectrogram reconstruction auxiliary task helped much in generalizing to unheard sounds on the Replica dataset as shown in Table \ref{tab:aux}, we did not observe the same improvement on the same setup for the Matterport3D dataset. The results further show a drop in performance on the heard sounds for the two datasets. Accordingly, we excluded this auxiliary task from the final architecture.
\input{tables/aux}

\subsection{Continuous Action Space Discritization}
\input{tables/dicretization_algo}
There are several challenges that stand against deploying these trained agents in real-life environments. One of these challenges is the discrete action space that the agent reason in. Nevertheless, several challenges stand as barriers to training the agent on continuous action space. 1. Habitat \cite{habitat19iccv} offers only a discrete set of actions. 2. SoundSpaces \cite{chen2020soundspaces} increases this hardship by providing the binaural room impulse response only for specific locations spaced with $0.5m$ in Replica dataset \cite{straub2019replica} and $1m$ in Matterport3D dataset \cite{chang2017matterport3d} as explained in Section \ref{soundspaces}, that eliminates any source of audio observations in any point other than the predefined points.  

We tried to tackle this problem by training on continuous action space to output velocity $v$ in the range between [0, 1] and angular velocity $\omega$ in the range between [-90, 90] then we calculated the resolution in $x$  and $y$ directions and added to the accumulated agent position in both directions finally, the agent moves to the grid location which has a higher value than the threshold in term of the difference between the accumulated agent position and the current agent position. We selected a threshold to be half the distance between the grid points ($0.25$ for Replica and $0.5$ Matterport3D). The planner executes the required actions to navigate towards the newly calculated grid position. We also added a new \textit{idle} action to the set of the discrete actions to enable the agent to take this action when the difference between the accumulated position and the current agent position is lower than the predefined threshold value so, the agent should stay in the same position as there is no audio observation in any middle position as explained before. This option is preferable to skipping the current step without taking any action to not lose the benefits of running different parallel environments steps simultaneously. The whole discretization algorithm is included in Algorithm \ref{alg:disc_algorithm}.

\subsection{Continuous Action Space Evaluation}
\begin{figure}[htpb!]
  \centering
    \includegraphics[width=0.9\textwidth,keepaspectratio]{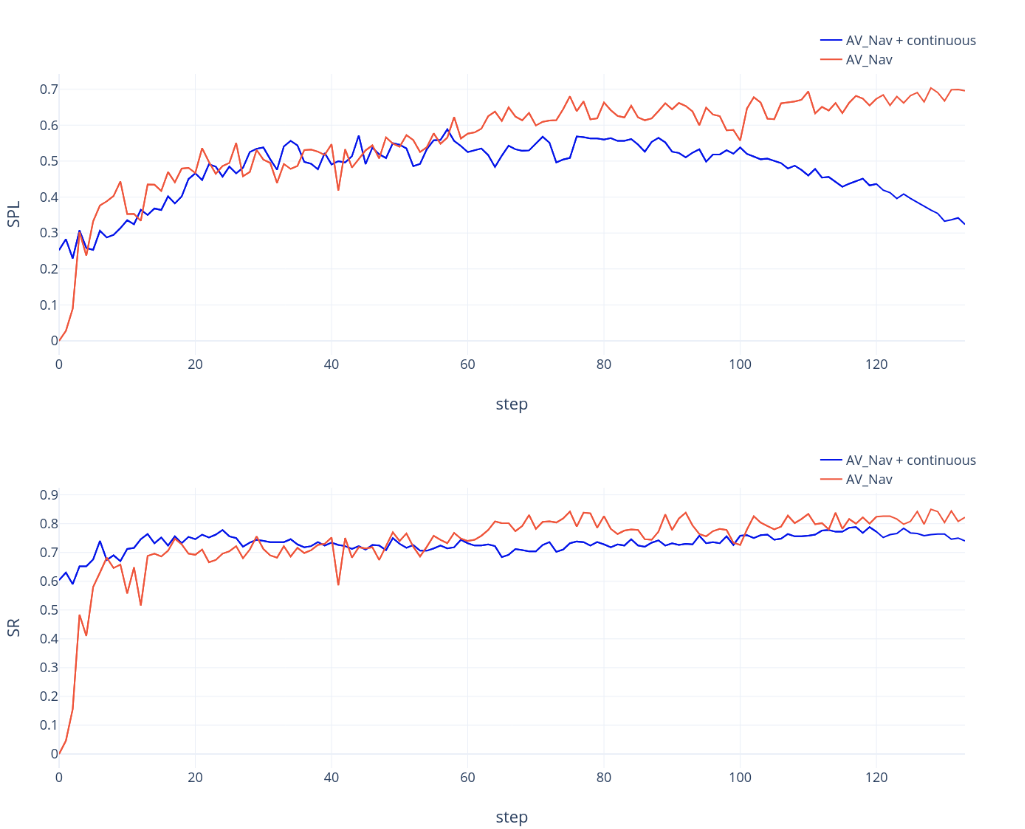}
    \caption{A comparison between the performance of AV-Nav baseline with discrete (red) and with continuous action space (blue). The metrics from top to bottom are SPL and SR. To ensure a fair comparison, SPL is calculated with respect to the discrete paths for both agents. }
    \label{fig:cont}
\end{figure}
Although we conducted the continuous action space experiments on the most straightforward task of learning a single sound on the Replica dataset, evaluating on the same sound, and taking the $stop$ action instead of the agent once the agent reaches the goal, the agent’s performance dropped significantly in the SPL metric. This demonstrates the complexity of learning to navigate using a continuous action space when the simulator’s set of actions is discrete, and the environment has limited discrete locations due to the limited availability of audio observations in other locations, as explained before in the SoundSpaces Section \ref{soundspaces}. The results of the comparison between the best continuous model and the AV-Nav baseline are shown in Figure \ref{fig:cont}. 

\section{Summary of Results}
We secured the first place in the \href{https://soundspaces.org/challenge}{SoundSpaces Challenge at The Conference on Computer Vision and Pattern Recognition (CVPR) 2021} by integrating the proposed audio complex scenarios Section \ref{sec:complex} and the spectrogram reconstruction Section \ref{sec:aux} as an auxiliary task with the AV-WaN architecture. This challenge was tackling the problem of generalizing to unheard sounds in Matterport3D \cite{chang2017matterport3d} on the static AudioGoal setup \cite{chen2020soundspaces}. Our approach outperformed the AV-WaN baseline by 11\ percentage points in terms of success weighted by path length (SPL) metric and 21 percentage points on the success rate (SR) metric on the standard test dataset, while on the unpublished challenge set dataset, this model increased the SPL by 13 percentage points and the SR by 18 percentage points.

On the evaluation of static AudioGoal task without complex scenarios, our new proposed architecture independently without integrating the complex scenarios improved the success rate by 12.5 percentage points on Replica \cite{straub2019replica} and  4 percentage points on Matterport3D on the clean unheard audio setup. While with integrating the audio complex scenarios, our proposed model increased the SPL by 26.7 percentage points on Replica and 15.4 percentage points on Matterport3D which outperfoms the challenge's winning model as well.

While on the evaluation of static AudioGoal task with complex scenarios, 
We trained the baselines on the complex audio scenarios as well to ensure a fair comparison. However, our model achieved a better SPL on the unheard sounds setup by 4 percentage points on Replica and 6.5 percentage points on Matterport3D. 

On the evaluation of the novel dynamic AudioGoal task without complex scenarios, our proposed architecture independently without integrating the complex audio scenarios achieved a success rate on the heard setup of 93.1\% on Replica and 95\% on Matterport3D. After adding the complex scenarios, our proposed architecture improved the success rate on the unheard sounds setup with 16.4 percentage points on Replica and 12.5 percentage points on Matterport3D. 

While on the evaluation of the dynamic AudioGoal task in the presence of the complex auudio scenarios, we also trained the baselines on the complex audio scenarios to to ensure a fair comparison. Nevertheless, our proposed approach outperforms the baseline on Matterport3D with an extra 13.2 percentage points on the unheard sounds setup. 

After reviewing the aforementioned results, we can conclude that 1. The novel dynamic audio-visual navigation task and the complex audio scenarios increase the complexity of the current task and cover a much broader range of scenarios. 2. Training in the presence of complex audio scenarios is beneficial for generalizing to unheard sounds and performing in noisy environments. 3. Our proposed model performs better on both tasks in clean and noisy environments. 

%% file: tables/static_without_complex.tex
\begin{table*}
  \footnotesize
  \centering
    \setlength{\tabcolsep}{1.5pt}
  \begin{tabular}{l|ccc|ccc|ccc|ccc}
    \toprule
    Model & \multicolumn{6}{c|}{Replica} & \multicolumn{6}{c}{MP3D} \\
    \cmidrule{2-13}
    & \multicolumn{3}{c|}{Multiple Heard} & \multicolumn{3}{c|}{Unheard} & \multicolumn{3}{c|}{Multiple Heard} & \multicolumn{3}{c}{Unheard} \\
    \cmidrule{2-13}
    & SPL & SR & SNA  & SPL & SR & SNA  & SPL & SR & SNA  & SPL & SR & SNA \\
    \midrule
    AV-Nav~\cite{chen2020soundspaces} & 54.1 & 73.9 & 30.3 & 34.1 & 51.1 & 16.7 & 53.7 & 69.8 & 30.2 & 28.3 & 38.2 & 14.8 \\ 
    AV-WaN~\cite{chen2020learning} &64.5   & \textbf{91.2} & 49.1 & 27.3 & 42.5 & 20.4 & 55.4 & 81.4 & 44.0 & 39.6 & 56.6 & 31.1 \\
    \textbf{Ours} & \textbf{71.9}& 85.9&\textbf{53.7} & \textbf{48.6} & \textbf{63.6} & \textbf{35.4} & \textbf{66.2}& \textbf{86.7}& \textbf{48.5}& \textbf{46.3}& \textbf{60.6}& \textbf{33.8}\\
    \midrule
    AV-Nav + complex &\textbf{63.3} &86.1 &32.2 & 44.3&63.4& 21.3&50.7 &70.3 &29.7 & 32.6 &48.8&17.2\\
    AV-WaN + complex & 62.9& \textbf{86.4} &\textbf{48.2} & 47.8& 73.7&36.2 &59.6 &\textbf{86.7} &45.5 & 47.4&69.4 &35.6 \\
    \textbf{Ours + complex} & 51.2 & 75.6 & 35.2 & \textbf{54.0} & \textbf{78.0} &\textbf{36.7} & \textbf{66.3}& 86.4& \textbf{47.7}& \textbf{55.0} & \textbf{73.0} & \textbf{38.6}  \\ 
    \midrule
    \textbf{Ours + complex + dynamic} & 46.9 & 60.8 & 34.8 & 37.4&52.7 &27.5& \textbf{66.4}& 84.1& \textbf{48.9}& 48.6& 65.3&35.6 \\
    \bottomrule
  \end{tabular}
  \vspace{-0.2cm}
  \caption{Results on the \textbf{static} Audio Goal task \textbf{without} complex scenarios. The heard experiments represent experiments trained on multiple sounds and evaluated on the same sounds but in unseen environments. The unheard sounds experiments represent experiments trained on multiple sounds and evaluated on multiple unheard sounds in unseen environments.}
  \label{tab:static}
  \vspace{-0.3cm}
\end{table*}

%% file: tables/static_with_complex.tex
\begin{table*}
  \footnotesize
  \centering
    \setlength{\tabcolsep}{2.5pt}
  \begin{tabular}{l|ccc|ccc|ccc|ccc}
    \toprule
    Model & \multicolumn{6}{c|}{Replica} & \multicolumn{6}{c}{MP3D} \\
    \cmidrule{2-13}
    & \multicolumn{3}{c|}{Multiple Heard} & \multicolumn{3}{c|}{Unheard} & \multicolumn{3}{c|}{Multiple Heard} & \multicolumn{3}{c}{Unheard} \\
    \cmidrule{2-13}
    & SPL & SR & SNA  & SPL & SR & SNA  & SPL & SR & SNA  & SPL & SR & SNA \\
    \midrule
    AV-Nav + complex &55.3 &\textbf{81.1} &27.5 & 43.8 & 65.1 & 21.5& 46.7& 69.4& 25.6&36.9&56.1&19.4\\
    AV-WaN + complex  &\textbf{54.0} &80.0 &\textbf{40.9} & 43.0& 66.5&32.3 & 57.6& 84.6 & 45.6&49.0 & 72.3 & 37.3  \\
    \textbf{Ours + complex} & 47.9 & 73.7 & 33.1 &  \textbf{47.8}& \textbf{73.4}& \textbf{33.3}& \textbf{64.1}& \textbf{86.4}& \textbf{46.8}& \textbf{55.5} & \textbf{76.0} & \textbf{40.4} \\
    \bottomrule
  \end{tabular}
  \vspace{-0.2cm}
  \caption{Results on the \textbf{static} Audio Goal task \textbf{with} complex scenarios. The heard experiments represent experiments trained on multiple sounds and evaluated on the same sounds but in unseen environments. The unheard sounds experiments represent experiments trained on multiple sounds and evaluated on multiple unheard sounds in unseen environments.}
  \label{tab:static_complex}
\end{table*}

%% file: tables/dynamic_without_complex.tex
\begin{table*}
  \footnotesize
  \centering
    \setlength{\tabcolsep}{1.0pt}
  \begin{tabular}{l|ccc|ccc|ccc|ccc}
    \toprule
    Model & \multicolumn{6}{c|}{Replica} & \multicolumn{6}{c}{MP3D} \\
    \cmidrule{2-13}
    & \multicolumn{3}{c|}{Multiple Heard} & \multicolumn{3}{c|}{Unheard} & \multicolumn{3}{c|}{Multiple Heard} & \multicolumn{3}{c}{Unheard} \\
    \cmidrule{2-13}
    & DSPL & SR & DSNA  & DSPL & SR & DSNA  & DSPL & SR & DSNA  & DSPL & SR & DSNA \\
    \midrule
    AV-Nav~\cite{chen2020soundspaces} & 44.0 &  69.5 & 14.6 & 22.3 & 35.0 & 7.5  & 39.9 & 70.1 & 16.4 & 21.8 & 38.7 & 8.8 \\ 
    AV-WaN~\cite{chen2020learning} &  56.6& 88.8 & 33.3 & \textbf{23.9} & \textbf{38.7} & \textbf{13.9} & 57.9 & 85.8 & 34.9 &  26.2 & 38.4 & 16.0 \\
    \textbf{Ours} & \textbf{62.4} & \textbf{93.1} & \textbf{33.6} & 23.1 & 36.0 & 12.4 & \textbf{63.9} & \textbf{95.0} & \textbf{36.7} & \textbf{29.9}   & \textbf{46.3} & \textbf{17.6}\\
    \midrule
    AV-Nav + complex& 43.1& 67.0& 14.6& 23.5& 38.4&7.5&37.7&67.3&15.5&20.4&35.5&8.4 \\
    AV-WaN + complex & \textbf{47.3}& \textbf{75.0} & \textbf{28.3}&29.4&51.6 & \textbf{17.6}& 55.4& 86.7& 33.8 &36.2 &61.5 & 22.1\\
    \textbf{Ours + complex} & 36.2 & 57.7 & 20.7 & \textbf{29.9} & \textbf{68.0} & 17.0 &  \textbf{61.6} &  \textbf{91.9} & \textbf{36.2} & \textbf{45.8} & \textbf{74.0} & \textbf{26.0}  \\ 
    \bottomrule
  \end{tabular}
  \vspace{-0.2cm}
  \caption{Results on the \textbf{dynamic} Audio Goal task \textbf{without} complex scenarios. The heard experiments represent experiments trained on multiple sounds and evaluated on the same sounds but in unseen environments. The unheard sounds experiments represent experiments trained on multiple sounds and evaluated on multiple unheard sounds in unseen environments.}
  \label{tab:dynamic}
  \vspace{-0.2cm}
\end{table*}

%% file: tables/dynamic_with_complex.tex
\begin{table*}
  \footnotesize
  \centering
    \setlength{\tabcolsep}{1.0pt}
  \begin{tabular}{l|ccc|ccc|ccc|ccc}
    \toprule
    Model & \multicolumn{6}{c|}{Replica} & \multicolumn{6}{c}{MP3D} \\
    \cmidrule{2-13}
    & \multicolumn{3}{c|}{Multiple Heard} & \multicolumn{3}{c|}{Unheard} & \multicolumn{3}{c|}{Multiple Heard} & \multicolumn{3}{c}{Unheard} \\
    \cmidrule{2-13}
    & DSPL & SR & DSNA  & DSPL & SR & DSNA  & DSPL & SR & DSNA  & DSPL & SR & DSNA \\
    \midrule
    AV-Nav + complex & 32.9&60.1&10.5 & 25.5& 45.2 &8.1 & 35.6 & 62.6 & 13.7 & 25.1 & 44.3 &9.7\\ 
    AV-WaN + complex & \textbf{44.0}&\textbf{66.1} &\textbf{25.6} & \textbf{34.2}&52.1 &\textbf{19.9} & 53.7 & 84.6 & 32.9 & 42.9 & 70.9 & 26.2\\
    \textbf{Ours + complex} & 32.6 & 55.3 & 18.8 & 27.5 &\textbf{52.6}  &16.1  & \textbf{59.9} & \textbf{92.8} & \textbf{33.3} & \textbf{52.1} & \textbf{84.1}& \textbf{30.4}  \\ 
    \bottomrule
  \end{tabular}
  \vspace{-0.2cm}
  \caption{Results on the \textbf{dynamic} Audio Goal task \textbf{with} complex scenarios. The heard experiments represent experiments trained on multiple sounds and evaluated on the same sounds but in unseen environments. The unheard sounds experiments represent experiments trained on multiple sounds and evaluated on multiple unheard sounds in unseen environments.}
  \label{tab:dynamic_complex}
\end{table*}

%% file: tables/augmentation_decomposition.tex
\begin{table*}

  \footnotesize
  \centering
    \setlength{\tabcolsep}{0.5pt}
  \begin{tabular}{l|ccc|ccc|ccc|ccc}
    \toprule
    Model & \multicolumn{6}{c|}{Replica} & \multicolumn{6}{c}{MP3D} \\
    \cmidrule{2-13}
    & \multicolumn{3}{c|}{Multiple Heard} & \multicolumn{3}{c|}{Unheard} & \multicolumn{3}{c|}{Multiple Heard} & \multicolumn{3}{c}{Unheard} \\
    \cmidrule{2-13}
    & SPL & SR & SNA  & SPL & SR & SNA  & SPL & SR & SNA  & SPL & SR & SNA \\
    \midrule
    Ours & \textbf{71.9}& \textbf{85.9}&\textbf{53.7} & 48.6 & 63.6 & 35.4 & 66.2& \textbf{86.7}& 48.5& 46.3& 60.6& 33.8\\
    Ours + second audio & 70.2 & 85.3 & 51.6 & \textbf{56.0} & 74.6 & \textbf{40.4} & 62.6 & 81.8 & 47.4 & 51.2 & 65.0 & 38.2 \\
    Ours + specaugment & 56.5 & 78.4 & 40.2 & 53.9 & 78.7 & 38.1 & 59.2 & 79.7 & 43.3 & 39.6 & 58.3 & 28.6 \\
    Ours + second audio + specaugment & 59.2 & 82.4 & 41.8 & 54.5 & \textbf{80.0} & 38.3 & 62.6 & 83.2 & \textbf{48.7} & 53.0 & \textbf{73.3} & \textbf{40.7} \\ 
    Ours + second audio + distractor & 61.5& 82.2 & 44.1& 48.7 & 72.1 & 34.6 & 64.7 & 83.3 & 47.8 & 52.0 & 70.1 & 37.8\\
    Ours + complex & 51.2 & 75.6 & 35.2 & 54.0 & 78.0 &36.7 & \textbf{66.3}& 86.4& 47.7& \textbf{55.0} & 73.0 & 38.6  \\ 
    \bottomrule
  \end{tabular}
  \caption{Decomposition of complex scenarios: Evaluation on the \textbf{static} Audio Goal task \textbf{without} complex scenarios.}
  \label{tab:decomp}
\end{table*}

\begin{table*}
  \footnotesize
  \centering
    \setlength{\tabcolsep}{0.5pt}
  \begin{tabular}{l|ccc|ccc|ccc|ccc}
    \toprule
    Model & \multicolumn{6}{c|}{Replica} & \multicolumn{6}{c}{MP3D} \\
    \cmidrule{2-13}
    & \multicolumn{3}{c|}{Multiple Heard} & \multicolumn{3}{c|}{Unheard} & \multicolumn{3}{c|}{Multiple Heard} & \multicolumn{3}{c}{Unheard} \\
    \cmidrule{2-13}
    & SPL & SR & SNA  & SPL & SR & SNA  & SPL & SR & SNA  & SPL & SR & SNA \\
    \midrule
    Ours& 24.5&31.7&18.0&20.0&26.9&15.0&40.0&51.3&28.7&32.1&43.1&23.4\\
    Ours + second audio & 31.0 & 41.7 & 22.6 & 23.6 & 33.1 & 16.8 & 46.9 & 62.1 & 35.8 & 43.1 & 56.5 & 32.8 \\
    Ours + specaugment & 47.1 & 67.7 & 33.3 & 45.7 & 67.8 & 32.3 & 50.7 & 70.9 & 36.5 & 40.4  & 58.0 & 28.7 \\
    Ours + second audio + specaugment & \textbf{49.3} & 70.4 & \textbf{34.9} & 46.1 & 68.8 & 32.6 & 55.6 & 79.0 & 43.3 & 53.4 & 72.9 & \textbf{40.7} \\ 
    Ours + second audio + distractor & 38.5& 56.1 & 27.6& 32.9 & 47.9 & 23.4 & 49.1 & 64.9 & 36.8 & 41.5 & 55.3 & 30.6\\
    Ours + complex & 47.9 & \textbf{73.7} & 33.1 &  \textbf{47.8} & \textbf{73.4} & \textbf{33.3} & \textbf{64.1} & \textbf{86.4} & \textbf{46.8}& \textbf{55.5} & \textbf{76.0} & 40.4 \\
    \bottomrule
  \end{tabular}
  \caption{Decomposition of complex scenarios: Evaluation on the \textbf{static} Audio Goal task \textbf{with} complex scenarios.}
  \label{tab:decomp_complex}
\end{table*}



%% file: tables/hpo.tex
\begin{table*}
  \footnotesize
  \centering
    \setlength{\tabcolsep}{1.5pt}
  \begin{tabular}{l|c}
  \toprule
Hyperparameter&Experimented \\\midrule
SpecAugment Component & Time Masking, Frequency Masking, \textbf{both} \\
Time \& Frequency Mask Parameters Replica & \textbf{(12F 32T)}(6F 16T)(27F 32T)(15F 32T)(12F 12T)\\
Time \& Frequency Mask Parameters MP3D & (12F 32T)(6F 16T)(27F 32T) (15F 32T)\textbf{(12F 12T)}
 \\
Reconstruction of spectrogram’s loss weight 
 & 1, 0.1, \textbf{0.01}, 0.001
 \\
Spectrogram auxiliary output
 & \textbf{current step}, next step, difference to last step  
 \\
Spectrogram reconstrruction architecture 
 & \textbf{Auto Encoder}, Variational Auto Encoder
 \\
Action Map Size 
 & (9x9), (5x5), \textbf{(3x3)} 
 \\
Complex Scenarios Components & distractors, second audio, specaugment, \textbf{combine all}
 \\

\bottomrule
  \end{tabular}
  \caption{A summary of the hyper parameter optimization experiments.}
  \label{tab:hpo}
\end{table*}

%% file: tables/hyperparameters.tex
\begin{table*}
  \footnotesize
  \centering
    \setlength{\tabcolsep}{2.5pt}
  \begin{tabular}{l|c}
  \toprule
Hyperparameter&Value \\\midrule
clip param & 0.1 \\
ppo epoch & 4 \\
num mini batch & 1 \\
value loss coef & 0.5 \\
entropy coef (AV-Nav) & 0.02 (0.2) \\
lr & $\expnumber{2.5}{-4}$ \\
eps & $\expnumber{1}{-5}$ \\
max grad norm & 0.5 \\
optimizer & Adam \\

steps number & 150 \\
gru hidden size& 512 \\
use gae& True \\
gamma& 0.99 \\
tau& 0.95 \\
linear clip decay&True \\
linear lr decay& True \\
exponential lr decay& False \\
exp decay lambda& 5.0 \\
reward window size& 50 \\
\midrule
number of processes (AV-Nav Matterport3d) & 5 (10) \\
number of updates (AV-Nav) & 10,000 (40,000)\\
\midrule
time mask param (Replica) & 32\\
frequency mask param (Replica) & 12\\
time mask param (Matterport3d) & 12\\
frequency mask param (Matterport3d) &12\\
\bottomrule
  \end{tabular}
  \caption{Hyperparameters used for training our model and the baselines. Differences across models are shown in parentheses.}
  \label{tab:hyper}
\end{table*}

%% file: tables/aux.tex
\begin{table*}
  \footnotesize
  \centering
    \setlength{\tabcolsep}{2.5pt}
  \begin{tabular}{l|ccc|ccc|ccc|ccc}
    \toprule
    Model & \multicolumn{6}{c|}{Replica} & \multicolumn{6}{c}{MP3D} \\
    \cmidrule{2-13}
    & \multicolumn{3}{c|}{Multiple Heard} & \multicolumn{3}{c|}{Unheard} & \multicolumn{3}{c|}{Multiple Heard} & \multicolumn{3}{c}{Unheard} \\
    \cmidrule{2-13}
    & SPL & SR & SNA  & SPL & SR & SNA  & SPL & SR & SNA  & SPL & SR & SNA \\
    \midrule
    Ours & \textbf{71.9}& \textbf{85.9} &\textbf{53.7} & 48.6 & 63.6 & 35.4 & \textbf{66.2}& \textbf{86.7}& \textbf{48.5}& 46.3& \textbf{60.6}& 33.8\\
    + specto aux & 62.8 & 85.4 & 45.6 & \textbf{59.2} & \textbf{85.2} & \textbf{42.9} & 63.5& 83.3&48 &\textbf{47.8} &58.4 &\textbf{34.15} \\

    \bottomrule
  \end{tabular}
  \vspace{-0.2cm}
  \caption{Comparison between the results of our model with and without spectrogram reconstruction as an auxiliary task on the \textbf{static} Audio Goal task \textbf{without} complex scenarios setup. The heard experiments represent experiments trained on multiple sounds and evaluated on the same sounds but in unseen environments. The unheard sounds experiments represent experiments trained on multiple sounds and evaluated on multiple unheard sounds in unseen environments.}
  \label{tab:aux}
  \vspace{-0.3cm}
\end{table*}

%% file: tables/dicretization_algo.tex
\begin{algorithm*}
\footnotesize
\textbf{Require: }{continuousActions: list of output continuous actions for all running environments, numEnvironments: number of running parallel environments previousIds: list of previous episodes' ids, currentIds: list of current episodes' ids, previousLocations: list of previous locations for all agents, currentLocations: list of current locations for all agents, intermidiateLocations: list of accumulated intermediate locations, discritizedLocations: list of calculated discritzed locations that the agent shall navigate to, gridResolution: distance between grid points, threshold: value that determines whether to move to next grid position or stay in the current one, threshold: values that decide whether to move or to stay at the same grid, noChangeCount: list of number of consecutive environments steps the agents did not change their locations on, deg2rad: function to convert the angle from degree to radian}
\newline
\SetAlgoLined

\For{i in range(numEnvironments)}
{
\If{currentIds[i] != previousIds[i] or currentLocations[i] != discritizedLocations[i]}
{
intermidiateLocations[i] = currentLocations[i] \\
discritizedLocations[i]  = currentLocations[i]
}
intermidiateLocations[i][0] +=  continuousActions[i][0] * $\cos{(deg2rad(continuousActions[i][1]))}$ \\
intermidiateLocations[i][1] +$=$  continuousActions[i][0] * $\sin{(deg2rad(continuousActions[i][1]))}$

\If{intermidiateLocations[i][0] >= currentLocations[i][0]}
{
xSteps = (intermidiateLocations[i][0] - currentLocations[i][0]) || gridResolution \\
xMod = (intermidiateLocations[i][0] - currentLocations[i][0]) \% gridResolution \\
discritizedLocations[i][0] = currentLocations[i][0] + xSteps * gridResolution \\
\If{xMod > threshold}
{
discritizedLocations[i][0] += gridResolution \\
}

}
\Else{
xSteps = (currentLocations[i][0]  - intermidiateLocations[i][0]) || gridResolution \\
xMod = (currentLocations[i][0] - intermidiateLocations[i][0]) \% gridResolution \\
discritizedLocations[i][0] = currentLocations[i][0] - xSteps * gridResolution \\
\If{xMod > threshold}
{
discritizedLocations[i][0] -= gridResolution \\
}
}

\If{intermidiateLocations[i][1] >= currentLocations[i][1]}
{
ySteps = (intermidiateLocations[i][1] - currentLocations[i][1]) || gridResolution \\
yMod = (intermidiateLocations[i][1] - currentLocations[i][1]) \% gridResolution \\
discritizedLocations[i][1] = currentLocations[i][1] + ySteps * gridResolution \\
\If{yMod > threshold}
{
discritizedLocations[i][1] += gridResolution \\
}

}
\Else{
ySteps = (currentLocations[i][1]  - intermidiateLocations[i][1]) || gridResolution \\
yMod = (currentLocations[i][1] - intermidiateLocations[i][1]) \% gridResolution \\
discritizedLocations[i][1] = currentLocations[i][1] - ySteps * gridResolution \\
\If{yMod > threshold}
{
discritizedLocations[i][1] -= gridResolution \\
}
}

}
previousIds = currentIds \\
\textbf{return} discritizedLocations
 \caption{Action Discritization Algorithm}
 \label{alg:disc_algorithm}
\end{algorithm*}

%% file: chapters/8-conclusions.tex
\chapter{Conclusions and Future Work}
\label{chap:conclusion}

We introduce the task of dynamic audio-visual navigation in unmapped complex 3D environments. We also integrate more complex real-world inspired audio scenarios with second audio sources, distractors, and spectrograms augmentation to increase the robustness of the embodied AI agents against noisy audio environments and enhance the generalization to unheard sounds. We propose a new architecture that fuses spatial geometrical occupancy maps with the binaural sound spectrogram to learn a robust navigation policy for the two tasks of static and dynamic audio-visual navigation. Our approach consistently outperforms the current state-of-the-art by a substantial margin on both benchmarks for unheard sounds on the Matterport3D and Replica datasets. We won the \href{https://soundspaces.org/challenge}{SoundSpaces Challenge at The Conference on Computer Vision and Pattern Recognition (CVPR) 2021} and submitted our main contributions in a paper with the title "Catch Me If You Hear Me: Dynamical Audio-Visual Navigation in Unmapped Complex 3D Environments with Moving Sounds" \cite{younes2021catch}.

Nonetheless, the sound sources in our setups have no embodiment shapes, so there is no semantic meaning for them; therefore, the agent will struggle to reach the sound-emitting source if the sound stops earlier. Consequently, we are interested in extending our novel benchmark in the future to include the semantics of the sound sources to include semantic audio-visual navigation tasks as well. Although we increased the complexity of audio scenarios to cover more real-world-like scenarios, we still did not test it in real-world environments to judge whether the added scenarios are enough for the agent to generalize to real-world settings or we still need to add more scenarios. Therefore, we are also interested in deploying our model in a real-world environment to test the efficiency of our randomized complex audio scenarios and update them accordingly to close any gap to the real-world settings. In addition to the limitations mentioned above, the SoundSpaces audio simulator provides the binaural room impulse responses for predefined locations in the environments, prohibiting the motion to any intermediate locations or receiving a continuous audio signal. It also limits the injection of any actuation noise or the use of continuous action space.